\pdfoutput=1
\documentclass[11pt]{article}

\usepackage{naacl2021}

\usepackage{times}
\usepackage{latexsym}

\usepackage{color}
\usepackage{amsmath}
\usepackage{bm}
\usepackage{booktabs}
\usepackage{multirow}
\usepackage{array}
\usepackage{enumitem}

\usepackage{graphicx}
\usepackage{mathtools, cuted}
\usepackage{dsfont}
\usepackage{hyperref}
\usepackage{amssymb}
\usepackage{amsfonts}
\usepackage{subcaption}
\usepackage{diagbox}
\usepackage[ruled,vlined,noresetcount]{algorithm2e}

\usepackage{xcolor}

\SetCommentSty{mycommfont}
\usepackage[T1]{fontenc}
\usepackage[utf8]{inputenc}

\usepackage{microtype}
\newcommand{\what}[1]{\textcolor{black}{#1}}
\DeclareMathOperator*{\argmax}{arg\,max}
\title{Active$^2$ Learning: Actively reducing redundancies in Active Learning methods for Sequence Tagging and Machine Translation}

\author{
  Rishi Hazra, Parag Dutta, Shubham Gupta,\\ 
  \textbf{Mohammed Abdul Qaathir, Ambedkar Dukkipati} \\
  \{\texttt{rishihazra, paragdutta, shubhamg,}\\ \texttt{mohammedq, ambedkar}\}\texttt{@iisc.ac.in} \\
  Indian Institute of Science, Bangalore}

\begin{document}
\maketitle
\begin{abstract}
While deep learning is a powerful tool for natural language processing (NLP) problems, successful solutions to these problems rely heavily on large amounts of annotated samples. However, manually annotating data is expensive and time-consuming. Active Learning (AL) strategies reduce the need for huge volumes of labeled data by iteratively selecting a small number of examples for manual annotation based on their estimated utility in training the given model. In this paper, we argue that since AL strategies choose examples independently, they may potentially select similar examples, all of which may not contribute significantly to the learning process. Our proposed approach, Active$\mathbf{^2}$ Learning (A$\mathbf{^2}$L), actively adapts to the deep learning model being trained to eliminate such redundant examples chosen by an AL strategy. We show that A$\mathbf{^2}$L is widely applicable by using it in conjunction with several different AL strategies and NLP tasks. We empirically demonstrate that the proposed approach is further able to reduce the data requirements of state-of-the-art AL strategies by $\approx \mathbf{3-25\%}$ on an absolute scale on multiple NLP tasks while achieving the same performance with virtually no additional computation overhead.
\end{abstract}

\section{Introduction}

\label{section:introduction}

\footnotetext[0]{In proceedings of NAACL-HLT 2021}

Active Learning (AL) \cite{Freund:1997:SSU:263100.263123,McCallum:1998:EEP:645527.757765} reduces the need for large quantities of labeled data by intelligently selecting unlabeled examples for expert annotation in an iterative process. Many Natural Language Processing (NLP) tasks like sequence tagging (NER, POS) and Neural Machine Translation (NMT) are very data-intensive and require a meticulous, time-consuming, and costly annotation process. On the other hand, unlabeled data is practically unlimited. Due to this, many researchers have explored applications of active learning for NLP \cite{10.5555/645528.657614,active_learning_clinical_survey}. A general AL method proceeds as follows: \textbf{(i)} The partially trained model for a given task is used to (possibly incorrectly) annotate the unlabeled examples. \textbf{(ii)} An \textit{active learning strategy} selects a subset of the newly labeled examples via a criterion that quantifies the perceived utility of examples in training the model. \textbf{(iii)} The experts verify/improve the annotations for the selected examples. \textbf{(iv)} These examples are added to the training set, and the process repeats. AL strategies differ in the criterion used in step (ii).

We claim that \textit{all} AL strategies select redundant examples in step \textbf{(ii)}. If one example satisfies the selection criterion, then many other \textit{similar} examples will also satisfy it (see the next paragraph for details). As the examples are selected independently, AL strategies redundantly choose all of these examples even though, in practice, it is enough to label only a few of them (ideally just one) for training the model. This leads to higher annotation costs, wastage of resources, and reduces the effectiveness of AL strategies. This paper addresses this problem by proposing a new approach called A$\mathbf{^2}$L (read as active-squared learning) that further reduces the redundancies of existing AL strategies.

Any approach for eliminating redundant examples must have the following qualities: \textbf{(i)} The redundancy should be evaluated in the context of the trained model. \textbf{(ii)} The approach should apply to a wide variety of commonly used models in NLP. \textbf{(iii)} It should be compatible with several existing AL strategies. The first point merits more explanation. As a model is trained, depending on the downstream task, it learns to focus on certain properties of the input. Examples that share these properties (for instance, the sentence structure) are similar from the model's perspective. If the model is confused about one such example, it will likely be confused about all of them. We refer to a similarity measure that is computed in the context of a model as a \textit{model-aware similarity} (Section \ref{section:modelawaresimilarity}).

\textbf{Contributions:} \textbf{(i)} We propose a Siamese twin-\cite{NIPS1993_769,Mueller:2016:SRA:3016100.3016291} based method for computing model-aware similarity to eliminate redundant examples chosen by an AL strategy. This Siamese network actively adapts itself to the underlying model as the training progresses. We then use clustering based on similarity scores to eliminate redundant examples. \textbf{(ii)} We develop a second, computationally more efficient approach that approximates the first one with a minimal drop in performance by avoiding the clustering step. Both of these approaches have the desirable properties mentioned above. \textbf{(iii)} We experiment with several AL strategies and NLP tasks to empirically demonstrate that our approaches are widely applicable and significantly reduce the data requirements of existing AL strategies while achieving the same performance. To the best of our knowledge, we are the first to identify the importance of model-aware similarity and exploit it to address the problem of redundancy in AL.  

\section{Related Work}
\label{section:related_work}
Active learning has a long and successful history in the field of machine learning \cite{Dasgupta:2009:APA:1577069.1577080,Awasthi:2017:PLE:3038256.3006384}. However, as the learning models have become more complex, especially with the advent of deep learning, the known theoretical results for active learning are no longer applicable \cite{W17-2630}. This has prompted a diverse range of heuristics to adapt the active learning framework to deep learning models \cite{W17-2630}. Many AL strategies have been proposed \cite{NIPS2006_3051,Haffari09activelearning,10.5555/1858681.1858769,Blundell:2015:WUN:3045118.3045290,Gal:2016:DBA:3045390.3045502}, however, since they choose the examples independently, the problem of redundancy (Section \ref{section:introduction}) applies to all. 

We experiment with various NLP tasks like named entity recognition (NER) \cite{ner-sekine2007}, part-of-speech tagging (POS) \cite{Marcus:1993:BLA:972470.972475}, neural machine translation (NMT) \cite{10.1007/978-3-540-30194-3_12,nepveu-etal-2004-adaptive,bahdanau2014neural, cho-etal-2014-properties,10.5555/2969033.2969173,ortiz-martinez-2016-online} and so on \cite{wordnet_landes,conll2000}. The tasks chosen by us form the backbone of many practical problems and are known to be computationally expensive during both training and inference. Many deep learning models have recently advanced the state-of-art for these tasks \cite{bahdanau2014neural,N16-1030,SiddhantEtAl:2018:Deep}. Our proposed approach is compatible with any NLP model, provided it supports the usage of an AL strategy.

Existing approaches have used model-independent similarity scores to promote diversity in the chosen examples. For instance, in \citet{Chen:2015:SAL:2868301.2868619}, the authors use cosine similarity to pre-calculate pairwise similarity between examples. We instead argue in favor of model-aware similarity scores and learn an expressive notion of similarity using neural networks. We compare our approach with a modified version of this baseline using cosine similarity on Infersent embeddings \cite{ConneauEtAl:2017:Supervised}.

%%%%%%%%%%%%%%%%%%%%%%%%%%%%%%%%%%%%%%%%%%%%%%%%%%%%%%%%%%%%%%%%%%%%%%%%%%%%%%%

\section{Proposed Approaches}
\label{section:proposed_approaches}

We use $\mathcal{M}$ to denote the model being trained for a given task. $\mathcal{M}$ has a module called encoder for encoding the input sentences. For instance, the encoder in $\mathcal{M}$ may be modeled by an LSTM \cite{Hochreiter:1997:LSM:1246443.1246450}. 

% ============================================================================ %

\subsection{Model-Aware Similarity Computation}
\label{section:modelawaresimilarity}

A measure of similarity between examples is required to discover redundancy. The simplest solution is to compute the cosine similarity between input sentences \cite{Chen:2015:SAL:2868301.2868619,W17-2630} using, for instance, the InferSent encodings \cite{ConneauEtAl:2017:Supervised}. However, sentences that have a low cosine similarity may still be similar in the context of the downstream task. Model $\mathcal{M}$ has no incentive to distinguish among such examples. A good strategy is to label a diverse set of sentences from the perspective of the model. For example, it is unnecessary to label sentences that use different verb forms but are otherwise similar if the task is agnostic to the tense of the sentence. A straightforward extension of cosine similarity to the encodings generated by model $\mathcal{M}$ achieves this. However, a simplistic approach like this would likely be incapable of discovering complex similarity patterns in the data. Next, we describe two approaches that use more expressive model-aware similarity measures. 

%================================================%

\begin{algorithm}[t]
\hspace{3pt}\KwData{$\mathcal{D}_1$: task dataset\; 
\hspace{29pt} $\mathcal{D}_2$: auxiliary similarity dataset}\
\KwIn{$\mathcal{D} \leftarrow 2\%$ of dataset $\mathcal{D}_1$\;
\hspace{20pt} $\mathrm{D} \longleftarrow \mathcal{D}_1 - \mathcal{D}$ \tcp*{unlabeled data}}\
\KwOut{Labeled data}
	\textbf{Initialization:} $\mathcal{D}$\;
    % $\mathcal{D} \longleftarrow$ \textsc{Annotate}($\mathcal{D}$)\;
    $\mathcal{M} \longleftarrow$ \textsc{Train}($\mathcal{D}$)\;
    $\mathcal{M}_{A^2L} \longleftarrow$
    \textsc{Train}($\mathcal{M}(\mathcal{D}_2)$)\;
	\For{$i\leftarrow 1$ \KwTo $l$}{
    $\mathcal{S} \leftarrow$ $\mathcal{AL}(\mathrm{D})$\tcp*{top 2\% confused samples}\
    \eIf{\tcp{Model-Aware Siamese}}
    {\For{each pair $(\mathrm{s}_{m},\mathrm{s}_{n})$ in $\mathcal{S}$}{\label{forins}
            $\mathbf{S}[m, n] \leftarrow \mathcal{M}_{A^2L}(\mathrm{s}_{m},\mathrm{s}_{n})$\;
         }
    $\mathcal{R} \longleftarrow$ \textsc{Cluster}($\mathbf{S}$)\;}
    {\tcp{Integrated Clustering} $\mathcal{R} \longleftarrow$ $\mathcal{M}_{A^2L}(\mathcal{S})$\;} 
     $\mathcal{R} \longleftarrow$ \textsc{Annotate}($\mathcal{R}$)\;
     $\mathcal{D} \longleftarrow \mathcal{D} \cup \mathcal{R}$\;
     $\mathcal{M} \longleftarrow$ \textsc{Retrain}($\mathcal{D}$)\
    }
%  }
\caption{Active$\mathbf{^2}$ Learning}
\label{alg:algorithm}
\end{algorithm}

% ============================================================================ 

\subsection{Model-Aware Siamese}
\label{section:modelawaresiamese}

In this approach, we use a Siamese twin's network \cite{NIPS1993_769} to compute the pairwise similarity between encodings obtained from model $\mathcal{M}$. A Siamese twin's network consists of an encoder (called the Siamese encoder) that feeds on the output of model $\mathcal{M}$'s encoder. The outputs of the Siamese encoder are used for computing the similarity between each pair of examples $a$ and $b$ as:
\begin{equation}
  \label{eq:similarity_computation}
  \mathrm{sim}(a, b) = \exp{(-\vert\vert \mathbf{o}^{a} - \mathbf{o}^{b}\vert\vert_2)},
\end{equation}
where $\mathbf{o}^a$ and $\mathbf{o}^b$ are the outputs of the Siamese encoder for sentences $a$ and $b$ respectively. Let $N$ denote the number of examples chosen by an AL strategy. We use the Siamese network to compute the entries of an $N \times N$ similarity matrix $\mathbf{S}$ where the entry $S_{ab} = \mathrm{sim}(a, b)$. We then use the spectral clustering algorithm \cite{NIPS2001_2092} on the similarity matrix $\mathbf{S}$ to group similar examples. A fixed number of examples from each cluster are added to the training dataset after annotation by experts.

We train the Siamese encoder to predict the similarity between sentences from the SICK (Sentences Involving Compositional Knowledge) dataset \cite{L14-1314} using mean squared error. This dataset contains pairs of sentences with manually annotated similarity scores. The sentences are encoded using the encoder in $\mathcal{M}$ and then passed on to the Siamese encoder for computing similarities. The encoder in $\mathcal{M}$ is kept fixed while training the Siamese encoder. The trained Siamese encoder is then used for computing similarity between sentences selected by an AL strategy for the given NLP task as described above. As $\mathcal{M}$ is trained over time, the distribution of its encoder output changes, and hence we periodically retrain the Siamese network to sustain its model-awareness. 

The number of clusters and the number of examples drawn from each cluster are user-specified hyper-parameters. The similarity computation can be done efficiently by computing the output of the Siamese encoder for all $N$ examples before evaluating equation \ref{eq:similarity_computation}, instead of running the Siamese encoder $O(N^2)$ times. The clustering algorithm runs in $O(N^3)$ time. For an AL strategy to be useful, it should select a small number of examples to benefit from interactive and intelligent labeling. We expect $N$ to be small for most practical problems, in which case the computational complexity added by our approach would only be a small fraction of the overall computational complexity of training the model with active learning (see Figure~\ref{bar_plots_comparison}).

\subsection{Integrated Clustering Model}
\label{section:clusteringviaclassification}

While the approach described in Section \ref{section:modelawaresiamese} works well for small to moderate values of $N$, it suffers from a computational bottleneck when $N$ is large. We integrate the clustering step into the similarity computation step to remedy this (see Figure~\ref{bar_plots_comparison}) and call the resultant approach as Integrated Clustering Model (\textit{Int Model}). Here, the output of model $\mathcal{M}$'s encoder is fed to a clustering neural network $\mathcal{C}$ that has $K$ output units with the softmax activation function. These units correspond to the $K$ clusters, and each example is directly assigned to one of the clusters based on the softmax output. 

To train the network $\mathcal{C}$, we choose a pair of similar examples (say $a$ and $b$) and randomly select a negative example (say $c$). We experimented with both SICK and Quora Pairs dataset\footnotemark[3]. All examples are encoded via the encoder of model $\mathcal{M}$ and then passed to network $\mathcal{C}$. The unit with the highest probability value for $a$ is treated as the ground-truth class for $b$. Minimizing the objective given below maximizes the probability of $b$ belonging to its ground truth class while minimizing the probability of $c$ belonging to the same class:
\begin{align}
  \label{eq:clustering_via_classification_loss}
  \mathcal{L}(a, b, c) = &-\lambda_1 \log p^b_{i_a} - \lambda_2 \log (1-p^c_{i_a}) \notag \\
  &+ \lambda_3 \sum_{k=1}^K p^b_k \log p^b_k.  
\end{align}
Here $\lambda_1$, $\lambda_2$, and $\lambda_3$ are user-specified hyperparameters, $p^x_j$ is the softmax output of the $j^{th}$ unit for example $x$, $j = 1, 2, \dots, K$, $x = a, b, c$, and $i_a = \argmax_{j \in \{1, 2, \dots K\}} p^a_j$. The third term encourages the utilization of all the K units across examples in the dataset. As before, a trained network $\mathcal{C}$ is used for clustering examples chosen by an AL strategy, and we select a fixed number of examples from each cluster for manual annotation.

It is important to note that: \textbf{(i)} These methods are not AL strategies. Rather, they can be used in conjunction with any existing AL strategy. Moreover, given a suitable Siamese encoder or clustering network $\mathcal{C}$, they apply to any model $\mathcal{M}$. \textbf{(ii)} Our methods compute model-aware similarity since the input to the Siamese or the clustering network is encoded using the model $\mathcal{M}$. The proposed networks also adapt to the underlying model as the training progresses. Algorithm \ref{alg:algorithm} describes our general approach called Active$^2$ Learning.
% of which these two approaches are specific instantiations.

%%%%%%%%%%%%%%%%%%%%%%%%%%%%%%%%%%%%%%%%%%%%%%%%%%%%%%%%%%%%%%%%%%%%%%%%%%%%%%%

\section{Experiments}
\label{section:experiments}

We establish the effectiveness of our approaches by demonstrating that they: \textbf{(i)} work well across a variety of NLP tasks and models, \textbf{(ii)} are compatible with several popular AL strategies,  and \textbf{(iii)} further reduce the data requirements of existing AL strategies, while achieving the same performance. In particular, we experiment\footnote[1]{Codes for the experiments are available at the following github link: \href{https://github.com/parag1604/A2L}{https://github.com/parag1604/A2L}.} with two broad categories of NLP tasks: (a) Sequence Tagging (b) Neural Machine Translation. Table \ref{table:datasets} lists these tasks and information about the corresponding datasets (including the two auxiliary datasets for training the Siamese network (Section \ref{section:modelawaresiamese})) used in our experiments. We begin by describing the AL strategies for the two kinds of NLP tasks.

%============================================%
%============================================%

\subsection{Active Learning Strategies for Sequence Tagging}
\label{section:activelearningstrategiesST}

\paragraph{Margin-based strategy: }Let $s({\mathbf{y}}) = \mathrm{P}_{\bm{\theta}}(\mathbf{Y} = \mathbf{y} \vert \mathbf{X} = \mathbf{x})$ be the score assigned by a model $\mathcal{M}$ with parameters $\bm{\theta}$ to output $\mathbf{y}$ for a given example $\mathbf{x}$. Margin is defined as the difference in scores obtained by the best scoring output $\mathbf{y}$ and the second best scoring output $\mathbf{y}{'}$, i.e.:
\begin{equation}
    \label{eq:marginbasedalstrategy}
    \mathrm{M}_{\mathrm{margin}} = \max_{\mathbf{\mathbf{y}}} s(\mathbf{y}) - \max_{\mathbf{y}^{'} \neq \mathbf{y}_\mathrm{max}} s(\mathbf{y}^{'}),
\end{equation}
where, $\mathbf{y}_\mathrm{max} = \argmax_\mathbf{y} s(\mathbf{y})$. The strategy selects examples for which $\mathrm{M}_{\mathrm{margin}} \leq \mathrm{\tau}_1$, where $\mathrm{\tau}_1$ is a hyper-parameter. We use Viterbi's algorithm \cite{Ryan:1993:VA:901051} to compute the scores $s(\mathbf{y})$.

% ============================================================================ %

\begin{table*}
    \small
    \centering
    \begin{tabular}{>{\centering\arraybackslash}m{1.3cm} >{\centering\arraybackslash}m{2.5cm} >{\centering\arraybackslash}m{1.9cm} m{8cm}}
        \toprule 
        \textbf{Task} & \textbf{Dataset} & \textbf{$\#$Train/$\#$Test} & \textbf{Example (Input/Output)}\\ 
        \midrule
        NER & CoNLL 2003 & \small{$14987$ / $3584$} & Fischler proposed EU measures after reports from Britain \linebreak 
        B-PER \,0 \,B-MISC \,0  \,0  \,0 \,0  \,B-LOC\\
        \hline
        POS & CoNLL 2003 & $14987$ / $3584$ & He ended the World Cup on the wrong note \linebreak
        PRP \, VBD \, DT \, NNP \, NNP \, IN \, DT \, JJ \, NN \\
        \hline
        CHUNK & CoNLL 2000 & $8936$ / $2012$ & The dollar posted gains in quiet trading \linebreak 
        B-NP \, I-NP \, B-VP \, B-NP \, B-PP \, B-NP \, I-NP\\
        \hline
        SEMTR & SEMCOR\footnotemark[2] & $13851$ / $4696$ & This section prevents the military departments \linebreak 
        0 \, Mental \, Agentive \, 0 \, 0 \, Object\\
        \hline
        NMT & Europarl (en $\to$ es) & $100000$ / $29155$ & (1) that is almost a personal record for me this autumn ! (2) es la mejor marca que he alcanzado este otono . \\
        \hline
        AUX & SICK & $9000 / 1000$ & (1) Two dogs are fighting. (2) Two dogs are wrestling and hugging. Similarity Score: 4 (out of 5)\\
        \hline
        AUX & Quora Pairs\footnotemark[3] & $16000$ / $1000$ (sets) \footnotemark[5] & (1) How do I make friends? (2) How to make friends? Label: 1 \\
        \bottomrule
    \end{tabular}
    \caption{Task and dataset descriptions. AUX is the task of training the Siamese network (Section \ref{section:modelawaresiamese}) or Integrated network $\mathcal{C}$ (Section \ref{section:clusteringviaclassification}). Citations: CoNLL 2003~\cite{article}, CoNLL 2000~\cite{conll2000}, SEMCOR\footnotemark[3], Europarl~\cite{koehn2005epc}, SICK~\cite{L14-1314}, Quora Pairs\footnotemark[4].}
    \label{table:datasets}
  \end{table*}

\paragraph{Entropy-based strategy:} All the NLP tasks that we consider require the model $\mathcal{M}$ to produce an output for each token in the sentence. Let $\mathbf{x}$ be an input sentence that contains $n(\mathbf{x})$ tokens and define $\bar{s}_j = \max_{o \in \mathcal{O}} \mathrm{P}_{\bm{\theta}}(y_j = o \vert \mathbf{X} = \mathbf{x})$ to be the probability of the most likely output for the $j^{th}$ token in $\mathbf{x}$. Here $\mathcal{O}$ is set of all possible outputs and $y_j$ is the output corresponding to the $j^{th}$ token in $\mathbf{x}$. We define the \textit{normalized entropy score} as:
\begin{equation}
    \label{eq:entropybasedalstrategy}
    \mathrm{M}_{\mathrm{entropy}} = -\frac{1}{n(\mathbf{x})} \sum_{j=1}^{n(\mathbf{x})} \bar{s}_j(\mathbf{y}) \log \bar{s}_j(\mathbf{y}).
\end{equation}

A length normalization $n(\mathbf{x})$ is added to avoid bias due to the example length as it may be undesirable to annotate longer length examples~\cite{claveau:hal-01621338}. The strategy selects examples with $\mathrm{M}_{\mathrm{entropy}} \geq \mathrm{\tau}_2$, where $\mathrm{\tau}_2$ is a hyper-parameter.

\paragraph{Bayesian Active Learning by Disagreement (BALD):} Due to stochasticity, models that use dropout \cite{Srivastava:2014:DSW:2627435.2670313} produce a different output each time they are executed. BALD \cite{HoulsbyEtAl:2011:BayesianActiveLearningForClassificationAndPreferenceLearning} exploits this variability in the predicted output to compute model uncertainty. Let $\mathbf{y}^{(t)}$ denote the best scoring output for $\mathbf{x}$ in the $t^{th}$ forward pass, and let $N$ be the number of forward passes with a fixed dropout rate, then:
\begin{equation}
  \label{eq:bald}
  \mathrm{M}_{\mathrm{bald}} = 1 - \frac{\mathrm{count}(\mathrm{mode}(\mathbf{y}^{(1)}, \dots, \mathbf{y}^{(N)}))}{N}.
\end{equation}
Here the $\mathrm{mode}(.)$ operation finds the output which is repeated most often among $\mathbf{y}^{(1)}, \dots, \mathbf{y}^{(N)}$, and the $\mathrm{count}(.)$ operation counts the number of times this output was encountered. This strategy selects examples with $\mathrm{M}_{\mathrm{bald}} \geq \mathrm{\tau}_3$ (hyper-parameter).

%%%%%%%%%%%%%%%%%%%%%%%%%%%%%%%%%%%%%%%%%%%%%%%%%%%%%%%%

\subsection{Active Learning Strategies for Neural Machine Translation\textsuperscript{2}}
\label{section:activelearningstrategiesNMT}

\footnotetext[2]{The AL strategies for NMT are ranking-based techniques, so we select the top 50\% of the candidates after sorting them in ascending (\ref{eq:LC},\ref{eq:CS}) or descending (\ref{eq:ADS}) order.}

\paragraph{Least Confidence (LC)}
This strategy estimates the uncertainty of a trained model on a source sentence $\mathbf{x}$ by calculating the conditional probability of the prediction $\hat{\mathbf{y}}$ conditioned on the source sentence\cite{Lewis94heterogeneousuncertainty}. 
\begin{equation}
  \label{eq:LC}
  \mathrm{M}_{\mathrm{LC}} = \frac{1}{\mathrm{n(\hat{y})}} \log \mathbf{P}(\hat{\mathbf{y}}|\mathbf{x})
\end{equation}
A length normalization of $\mathrm{n(\hat{y})}$ (length of the predicted translation $\hat{\mathbf{y}}$) is added.

\paragraph{Coverage Sampling (CS)}
% It is reasonable to expect the translation model to translate all the source tokens into appropriate target tokens. 
A translation model is said to \textit{cover} the source sentence if it translates all of its tokens. Coverage is estimated by mapping a particular source token to its appropriate target token, without which the model may suffer from under-translation or over-translation issues ~\cite{tu-etal-2016-modeling}.
% The attention mechanism in LSTM based encoder-decoder architecture can model word alignment between the source and the corresponding model translation.
\citet{peris-casacuberta-2018-active} proposed to use translation coverage as a measure of uncertainty by:
\begin{equation}
    \label{eq:CS}
    \mathrm{M}_{\mathrm{CS}} = \frac{\sum_{\mathrm{j}=1}^{n(\mathbf{x})}\log (\min(\sum_{\mathrm{i}=1}^{n(\mathbf{\hat{y}})}\alpha_{\mathrm{i},\mathrm{j}}, 1))}{n(\mathbf{x})}
\end{equation}
Here $\alpha_{\mathrm{i},\mathrm{j}}$ denotes the attention probability calculated by the model for the $j^{th}$ source word in predicting the $i^{th}$ target word. It can be noted that the coverage score will be $0$ for samples for which the model almost fully covers the source sentences.

\paragraph{Attention Distraction Sampling (ADS)}
\citet{peris-casacuberta-2018-active} claimed that in translating an uncertain sample, the model's attention mechanism would be \textit{distracted} (dispersed throughout the sentence). Such samples yield attention probability distribution with light tails (e.g., uniform distribution), which can be obtained by taking the Kurtosis of the attention weights for each target token $\mathbf{y}_\mathrm{i}$.
\begin{equation}
    \mathrm{Kurt}(\mathbf{y}_\mathrm{i}) = \frac{\frac{1}{n(\mathbf{x})}\sum_{\mathrm{j}=1}^{n(\mathbf{x})}(\alpha_{\mathrm{i},\mathrm{j}}-\frac{1}{n(\mathbf{x})})^4}{(\frac{1}{n(\mathbf{x})}(\sum_{\mathrm{j}=1}^{n(\mathbf{x})}\alpha_{\mathrm{i},\mathrm{j}}-\frac{1}{n(\mathbf{x})}))^2}
\end{equation}
where $\frac{1}{n(\mathbf{x})}$ is the mean of the distribution of the attention weights (for a target word) over the source words. The kurtosis value will be lower for distributions with light tails, so the average of the negative kurtosis values for all words in the target sentence is used as the distraction score.
\begin{equation}
  \label{eq:ADS}
  \mathrm{M}_{\mathrm{ADS}} = \frac{\sum_{\mathrm{i}=1}^{n(\mathbf{y})}-\mathrm{Kurt}(\mathbf{y}_\mathrm{i})}{n(\mathbf{y})}
\end{equation}

\footnotetext[3]{From a subset of the Brown Corpus \cite{doi:10.1177/007542428501800107}, using splits from ~\citet{martinez:alonso-plank-2017-multitask}}
\footnotetext[4]{\href{https://www.quora.com/q/quoradata/First-Quora-Dataset-Release-Question-Pairs}{https://www.quora.com/q/quoradata/First-Quora-Dataset-Release-Question-Pairs}}

% ============================================================================ %
\begin{figure}
\centering  
\includegraphics[width=\linewidth]{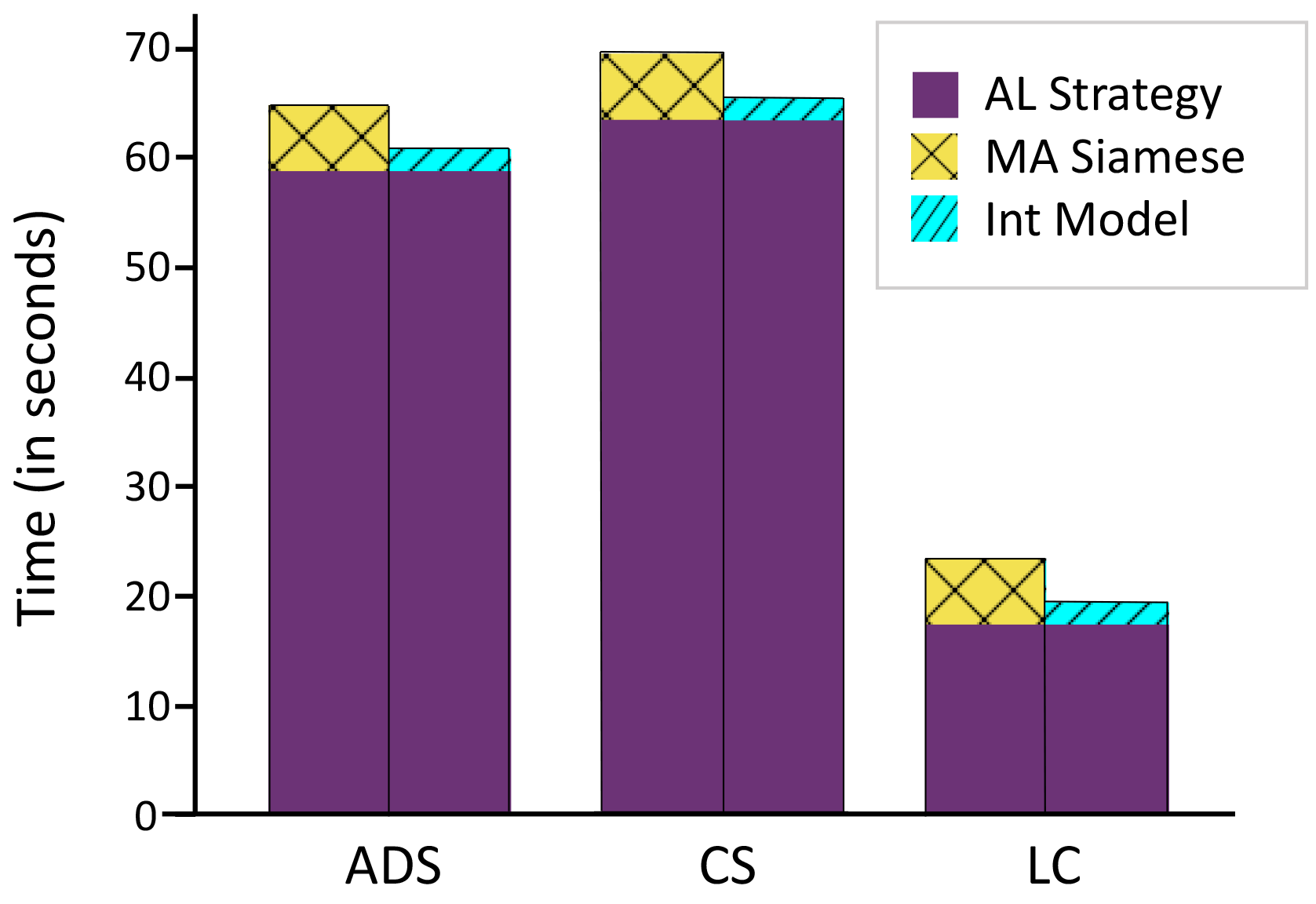}
\caption{Comparison of time taken for one data selection step in NMT task by the Model Aware (MA) Siamese and Integrated Clustering (Int) Model across different ALS. It can be observed that A$^2$L adds a negligible overhead ($\approx \frac{1}{12}$ of the time taken for ALS) to the overall process.}
\label{bar_plots_comparison}
\end{figure}

\begin{table}[h!]
	\small
	 \begin{center}
		 \begin{tabular}{ >{\centering\arraybackslash}m{1.0cm} >{\centering\arraybackslash}m{1.1cm} >{\centering\arraybackslash}m{1.7cm} >{\centering\arraybackslash}m{1.7cm}}
			 \toprule
			 \textbf{Task} & \textbf{Dataset} & \textbf{\% of train data used to reach $99 \%$ of full-data F-Score} & \textbf{\% less data required to reach $99 \%$ of full-data F-score}\\[0.4ex] 
			 \midrule
			 POS & CoNLL 2003 & 25\% & 16\%\\
			 \midrule
			 NER & CoNLL 2003 & 37\% & 3\%\\
			 \midrule
			 SEMTR & SEMCOR & 35\% & 25\%\\
			 \midrule
			 CHUNK & CoNLL 2000 & 23\% & 11\%\\
			 \bottomrule
		 \end{tabular}
	 \end{center}
	 \caption{Fraction of data used for reaching full dataset performance and the corresponding absolute percentage reduction in the data required over the None baseline that uses active learning strategy without the A$^\mathbf{2}$L step for the best AL strategy (BALD in all cases). Refer Fig~\ref{original_graphs} in Appendix for CHUNK plots.}
	 \label{tab_results}
	 \vspace{-0.55cm}
 \end{table}

\subsection{Details about Training}
\label{section:detailsabouttraining}

For sequence tagging, we use two kinds of architectures: CNN-BiLSTM-CRF model (CNN for character-level encoding and BiLSTM for word-level encoding) and a BiLSTM-BiLSTM-CRF model (BiLSTM for both character-level and word-level encoding) (\citet{N16-1030,SiddhantEtAl:2018:Deep}).
For the translation task, we use LSTM based encoder-decoder architecture with Bahdanau attention \cite{bahdanau2014neural}.
These models were chosen for their performance and ease of implementation.

The Siamese network used for model-aware similarity computation (Section \ref{section:modelawaresiamese}) consists of two bidirectional LSTM (BiLSTM) encoders. We pass each sentence in the pair from the SICK dataset to model $\mathcal{M}$ and feed the resulting encodings to the Siamese BiLSTM encoder. The output is a concatenation of terminal hidden states of the forward and backward LSTMs, which is used to compute the similarity score using (\ref{eq:similarity_computation}). As noted before, we keep model $\mathcal{M}$ fixed while training the Siamese encoders and use the trained Siamese encoders for computing similarity between examples chosen by an AL strategy. We maintain the model-awareness by retraining the Siamese after every $10$ iterations.

The architecture of the clustering model $\mathcal{C}$ (Section \ref{section:clusteringviaclassification}) is similar to that of the Siamese encoder. Additionally, it has a linear layer with a softmax activation function that maps the concatenation of terminal hidden states of the forward and backward LSTMs to $K$ units, where $K$ is the number of clusters. To assign an input example to a cluster, we first pass it through the encoder in $\mathcal{M}$ and feed the resulting encodings to the clustering model $\mathcal{C}$. The example is assigned to the cluster with the highest softmax output. 
% We train this network via the procedure described in Section \ref{section:clusteringviaclassification} using the Quora Pairs dataset. 
This network is also retrained after every $10$ iterations to retain model-awareness.

The initial data splits used for training the model $\mathcal{M}$ were set at $2\%$ of randomly sampled data for Sequence Tagging ($20\%$ for NMT). These are in accordance with the splitting techniques used in the existing literature on AL~\cite{SiddhantEtAl:2018:Deep, liu-etal-2018-learning}. The model is then used to provide input to train the Siamese/Clustering network using the SICK/Quora Pairs\footnotetext[5]{We process the dataset to use only those sentences which are present in at least 5 other pairs. We retrieve 16000 sets, each with a source sentence and 5 other samples (comprising both positive and negative labels). An additional 1000 sets were generated for evaluation.}. At each iteration, we gradually add another $2\%$ of data for sequence tagging ($5\%$ for NMT) by retrieving low confidence examples using an AL strategy, followed by clustering to extract the most representative examples. We average the results over five independent runs with randomly chosen initial splits. [Hyperparameters details in Appendix \ref{Table of hyperparameters}].

% ==================== %

\begin{figure*}
	\centering
	\begin{subfigure}[t]{0.3\textwidth}
		\centering
		\includegraphics[width=\linewidth]{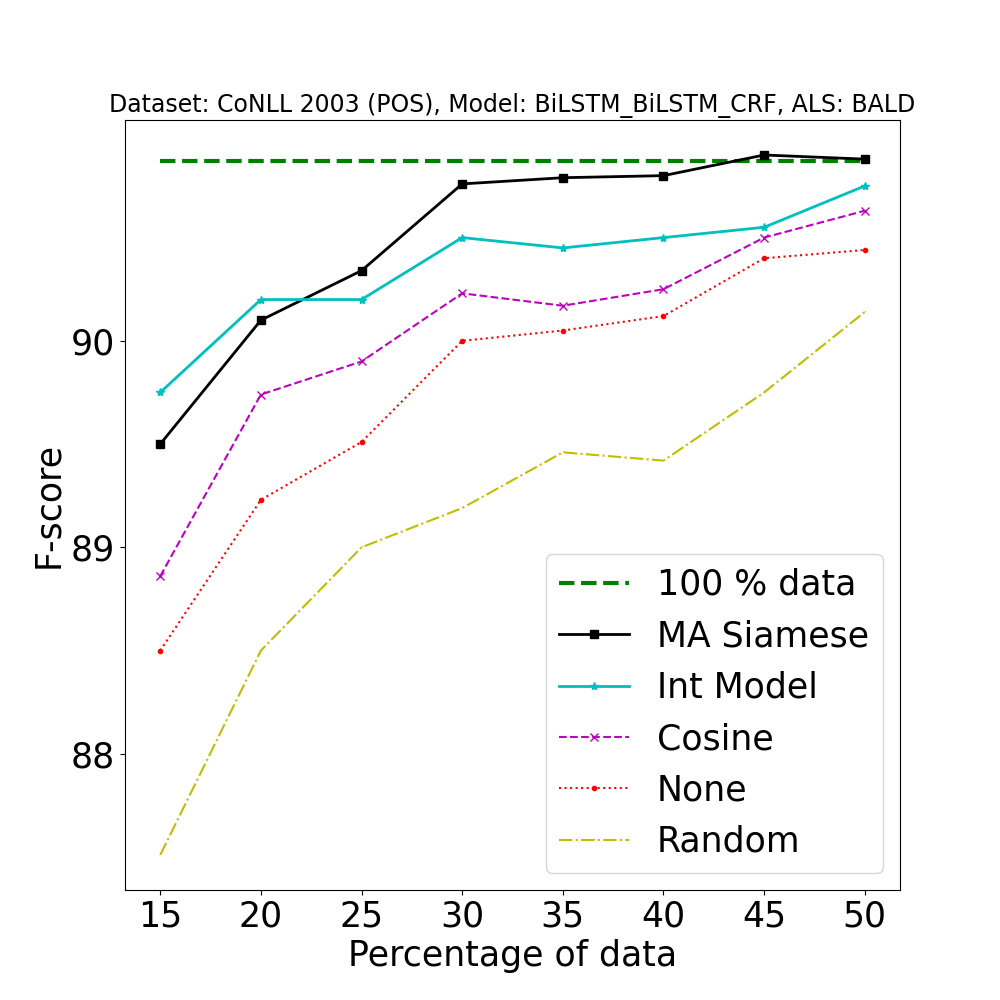}
		% \caption{Plot 4}
	\end{subfigure}%
	~ 
	\begin{subfigure}[t]{0.3\textwidth}
		\centering
		\includegraphics[width=\linewidth]{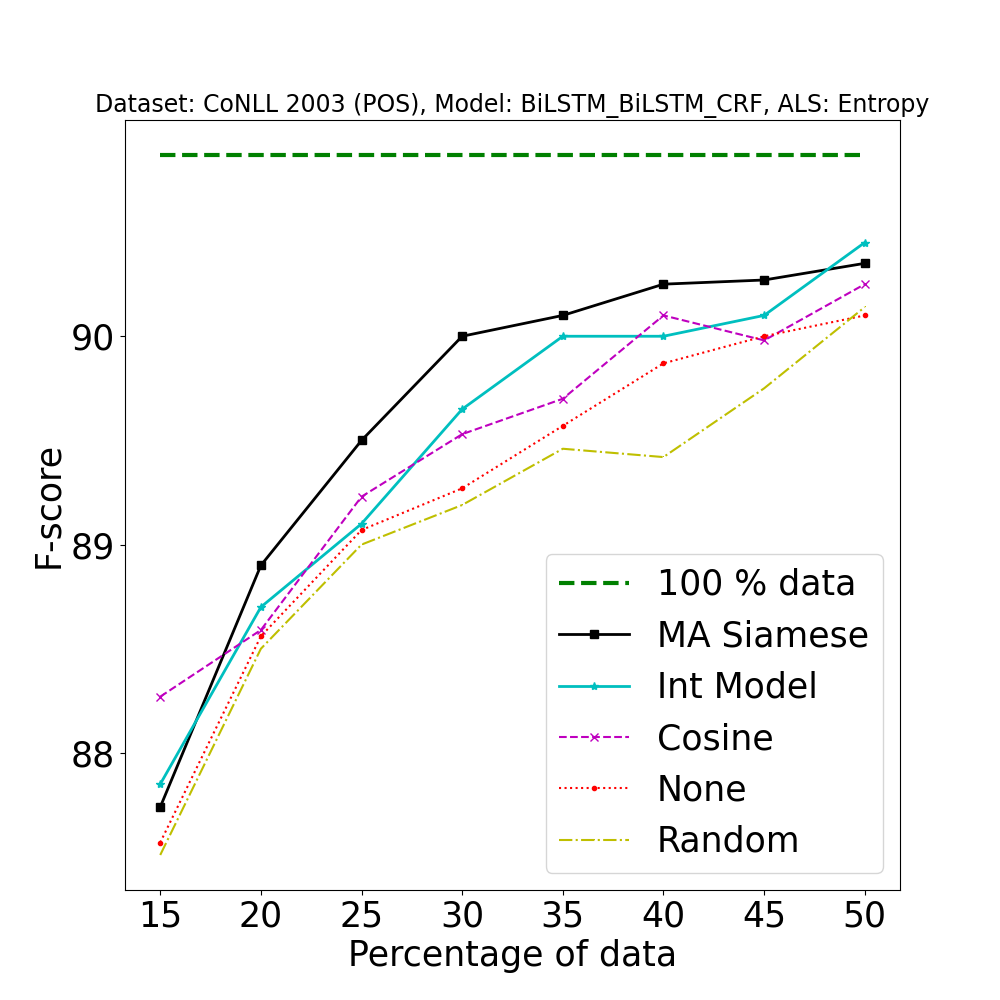}
		%\caption{Plot 5}
	\end{subfigure}%
	~
	\begin{subfigure}[t]{0.3\textwidth}
		\centering
		\includegraphics[width=\linewidth]{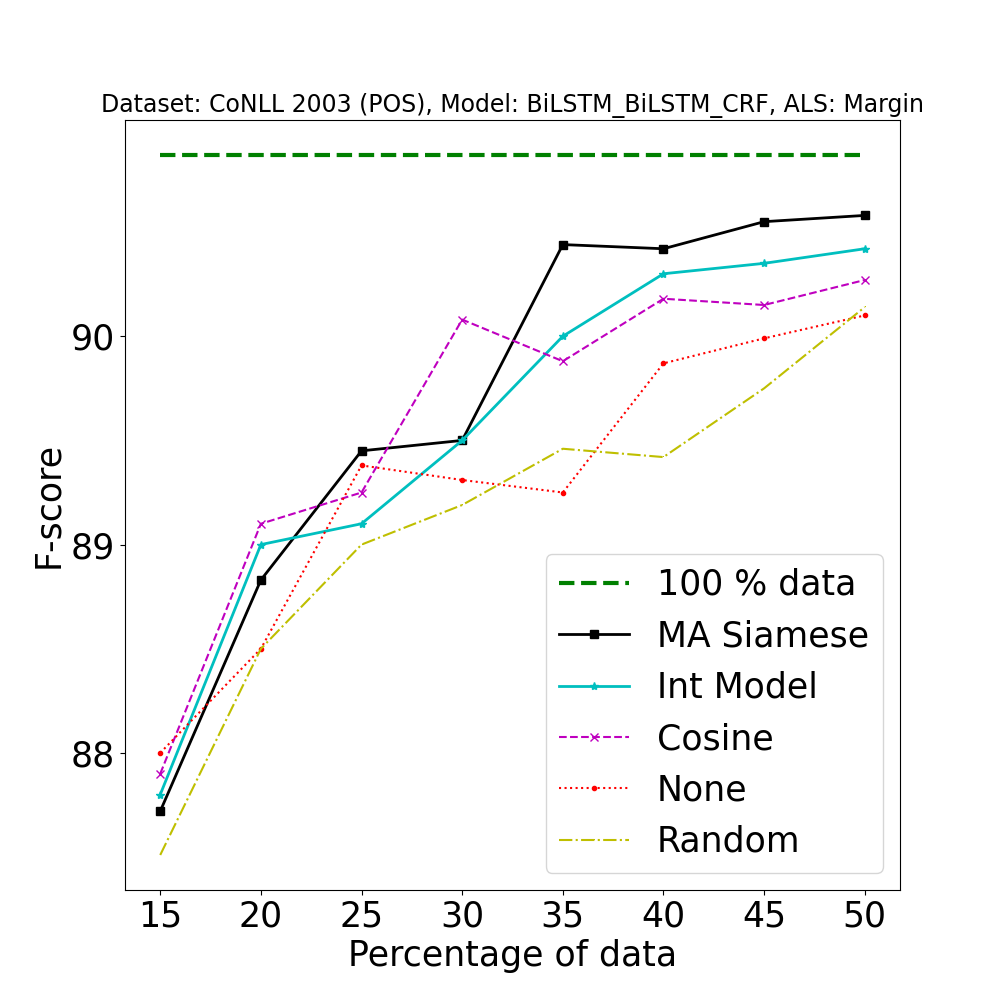}
		%\caption{Plot 6}
	\end{subfigure}%
	\\
	\begin{subfigure}[t]{0.3\textwidth}
		\centering
		\includegraphics[width=\linewidth]{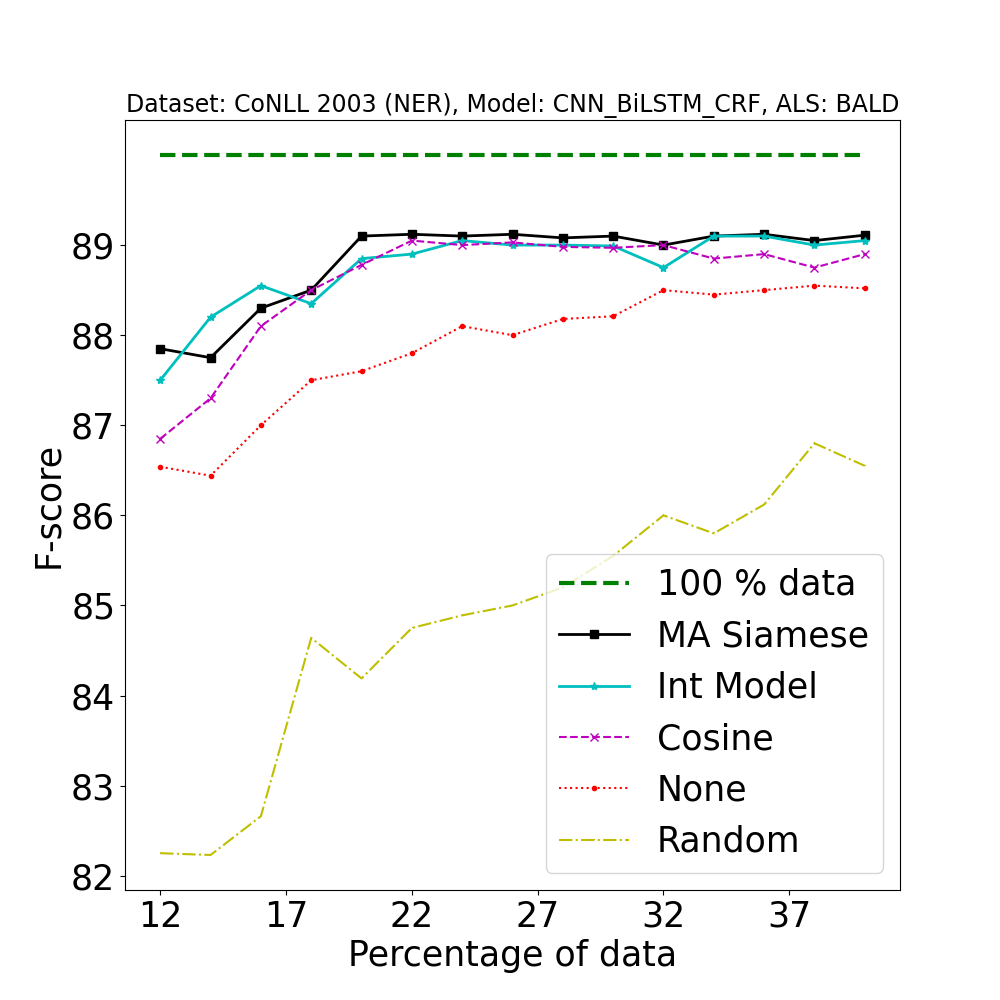}
		% \caption{Plot 4}
	\end{subfigure}%
	~ 
	\begin{subfigure}[t]{0.3\textwidth}
		\centering
		\includegraphics[width=\linewidth]{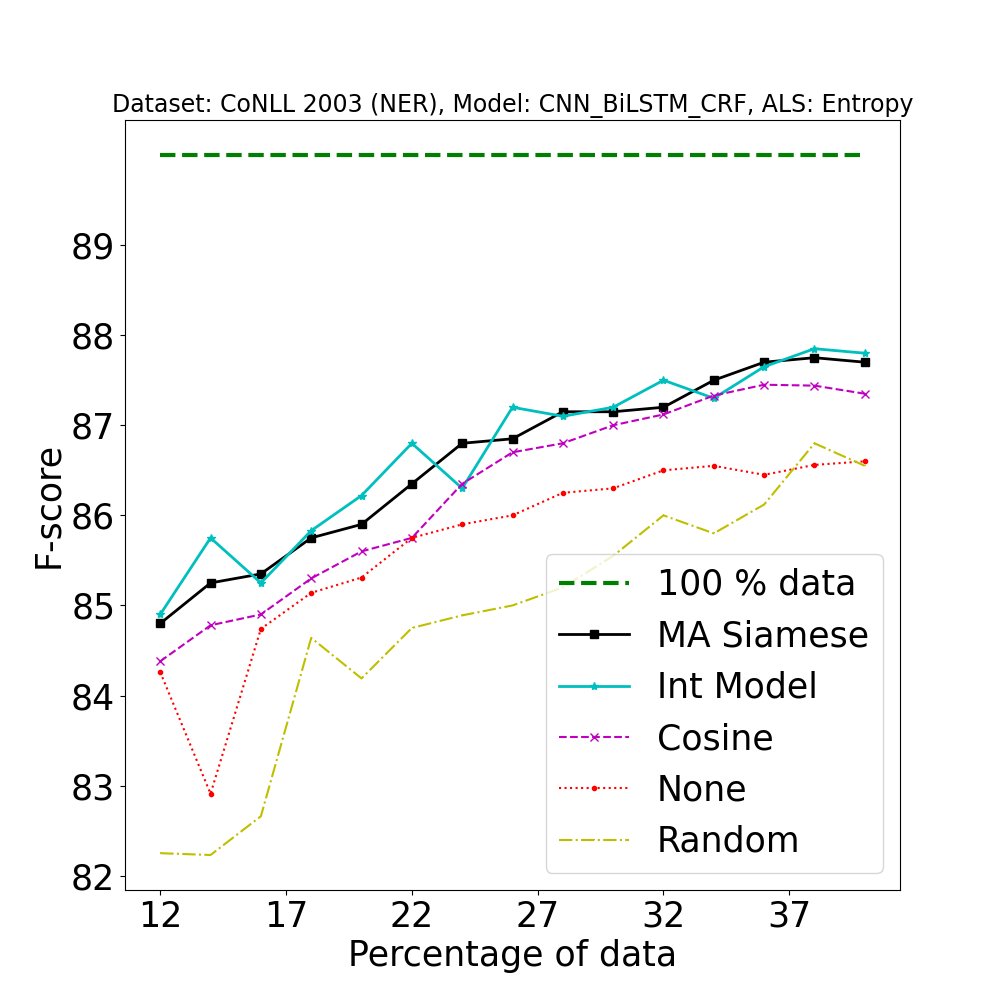}
		%\caption{Plot 5}
	\end{subfigure}%
	~
	\begin{subfigure}[t]{0.3\textwidth}
		\centering
		\includegraphics[width=\linewidth]{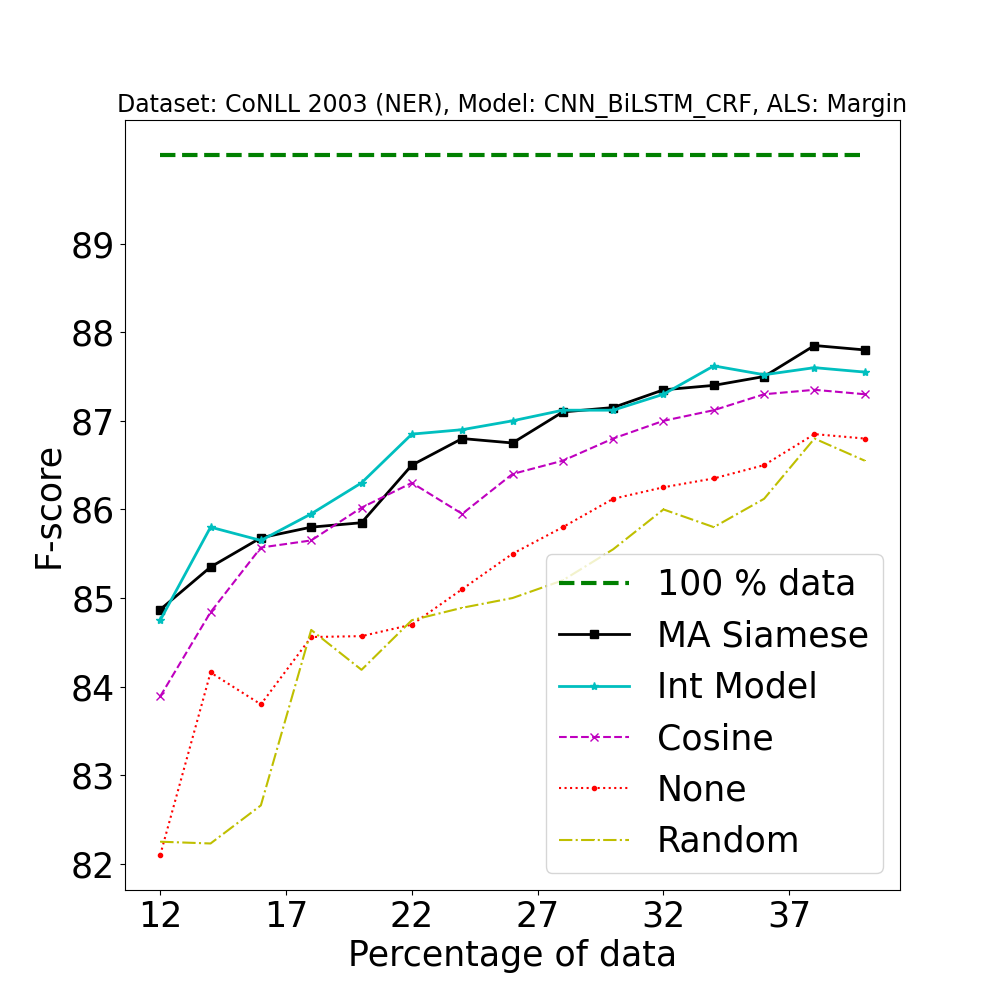}
		%\caption{Plot 6}
	\end{subfigure}%
	\\
	\begin{subfigure}[t]{0.3\textwidth}
		\centering
		\includegraphics[width=\linewidth]{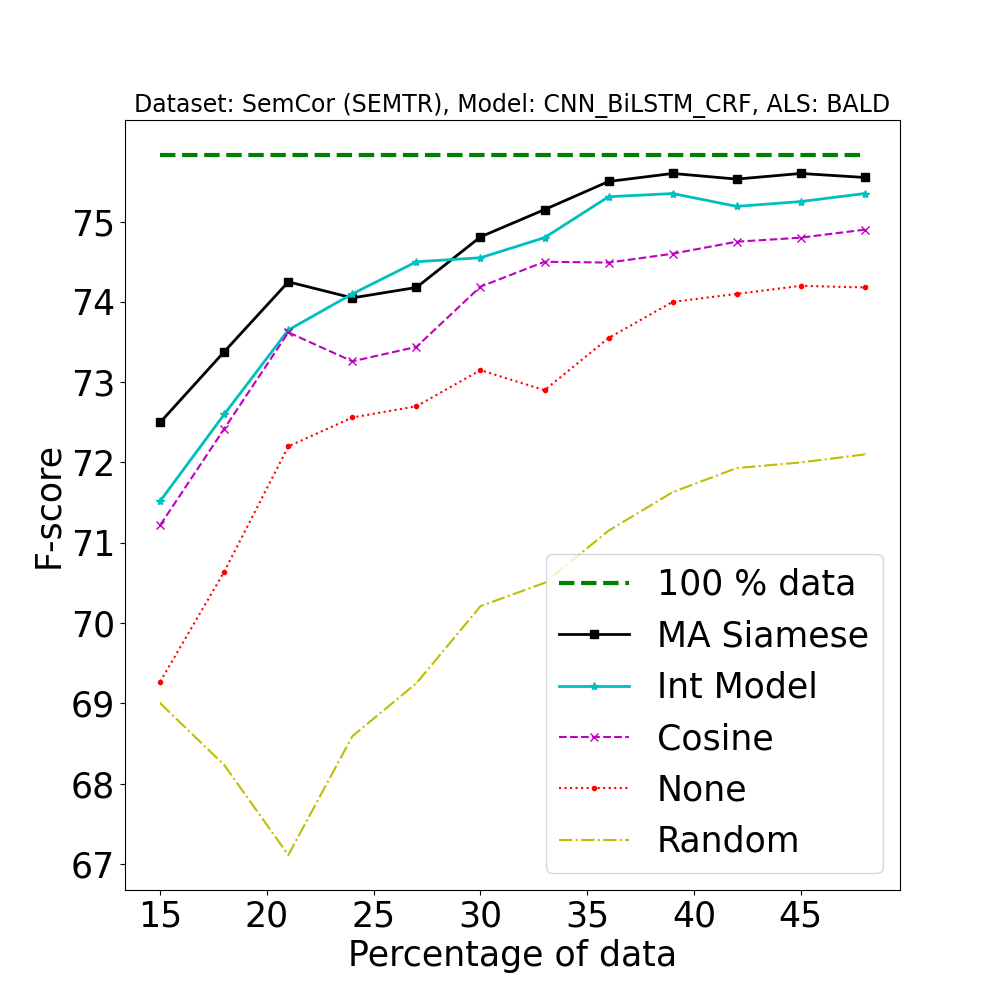}
		%\caption{Plot 1}
	\end{subfigure}%
	~ 
	\begin{subfigure}[t]{0.3\textwidth}
		\centering
		\includegraphics[width=\linewidth]{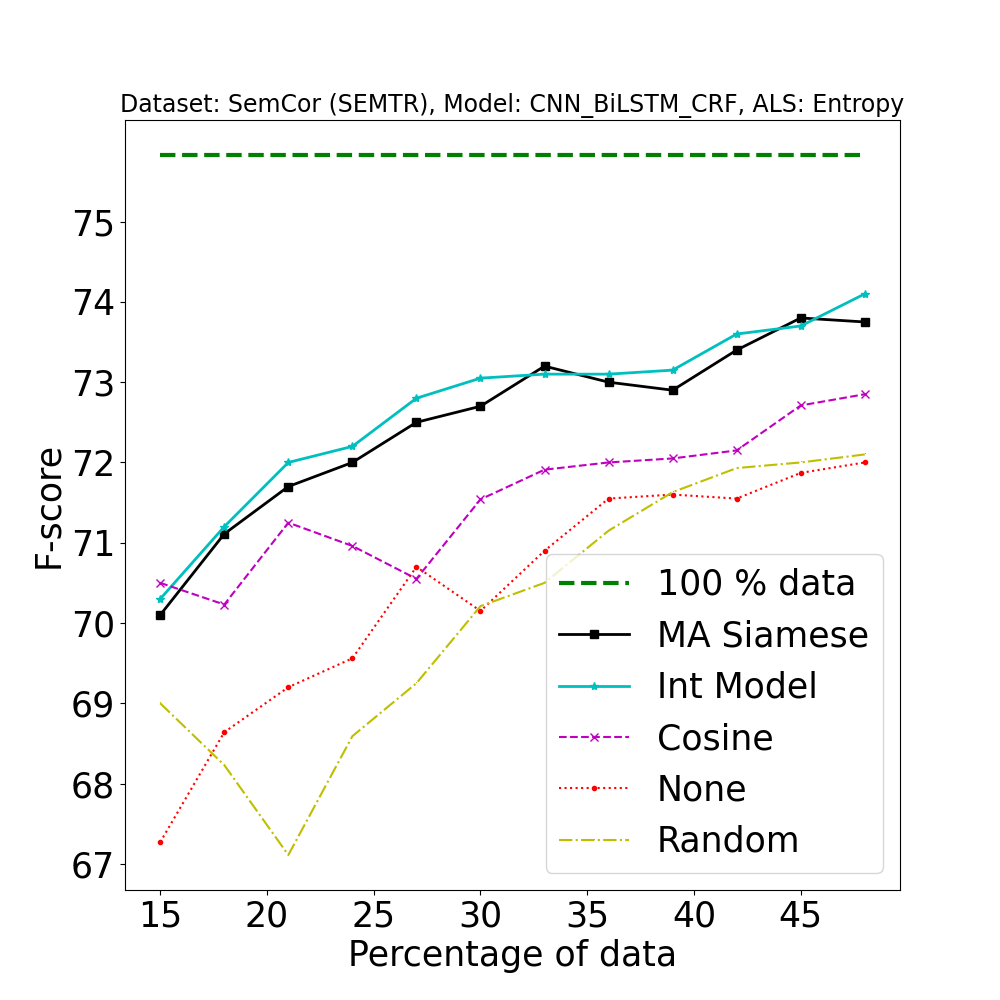}
		%\caption{Plot 2}
	\end{subfigure}%
	~
	\begin{subfigure}[t]{0.3\textwidth}
		\centering
		\includegraphics[width=\linewidth]{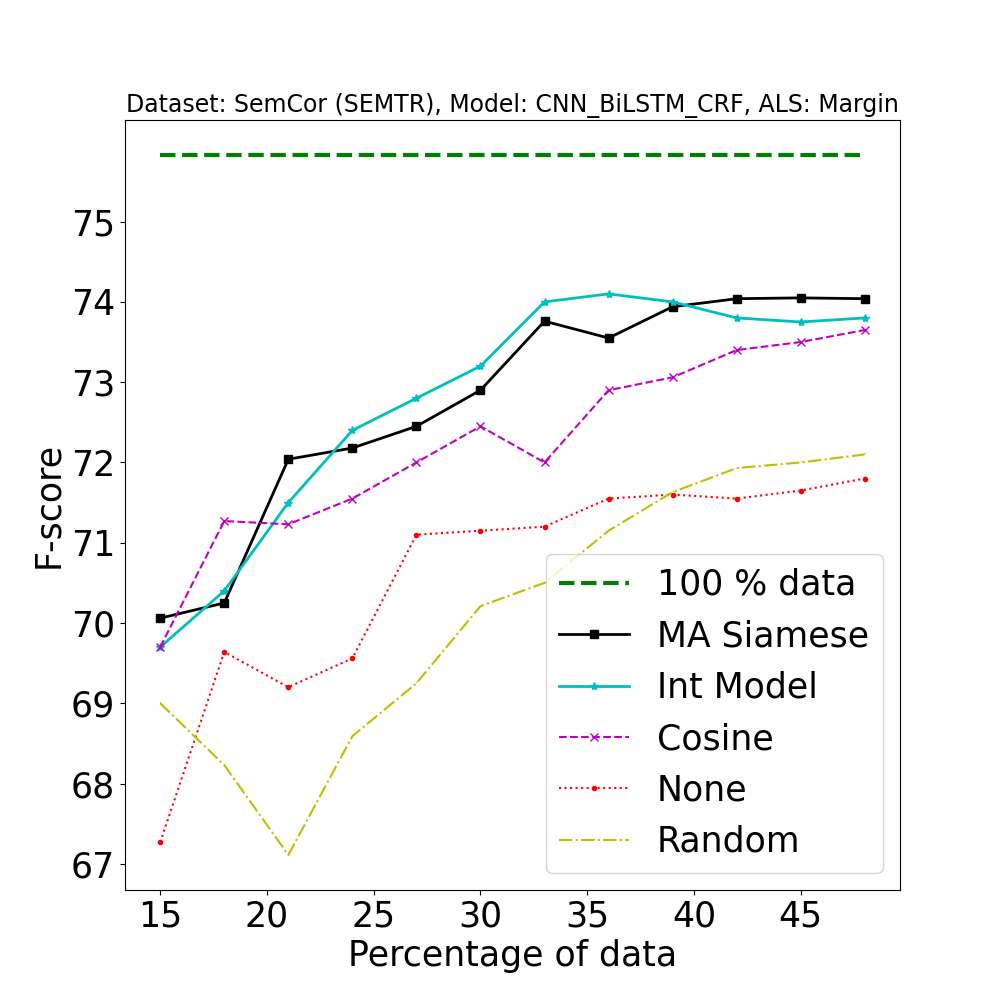}
		%\caption{Plot 3}
	\end{subfigure}%
	\\
	\begin{subfigure}[t]{0.3\textwidth}
		\centering
		\includegraphics[width=\linewidth]{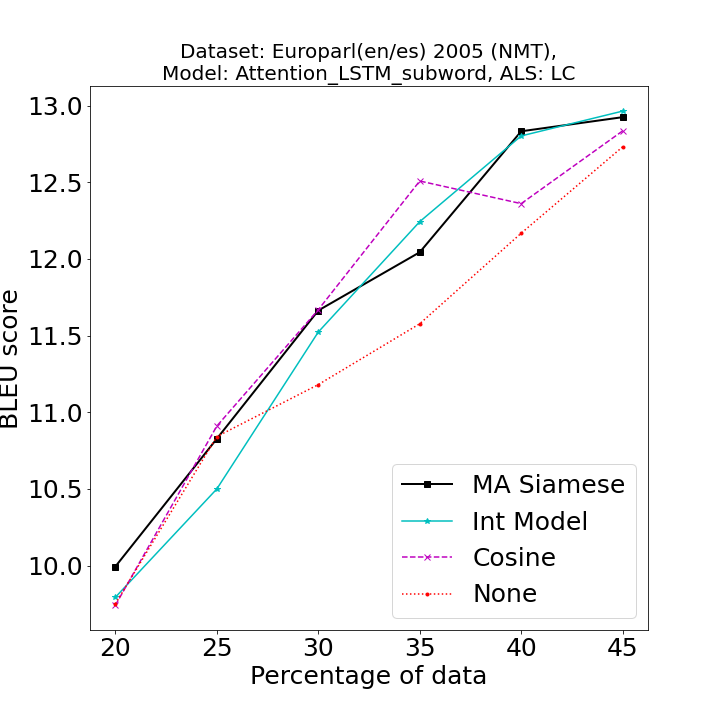}
		%\caption{Plot 10}
	\end{subfigure}%
	~ 
	\begin{subfigure}[t]{0.3\textwidth}
		\centering
		\includegraphics[width=\linewidth]{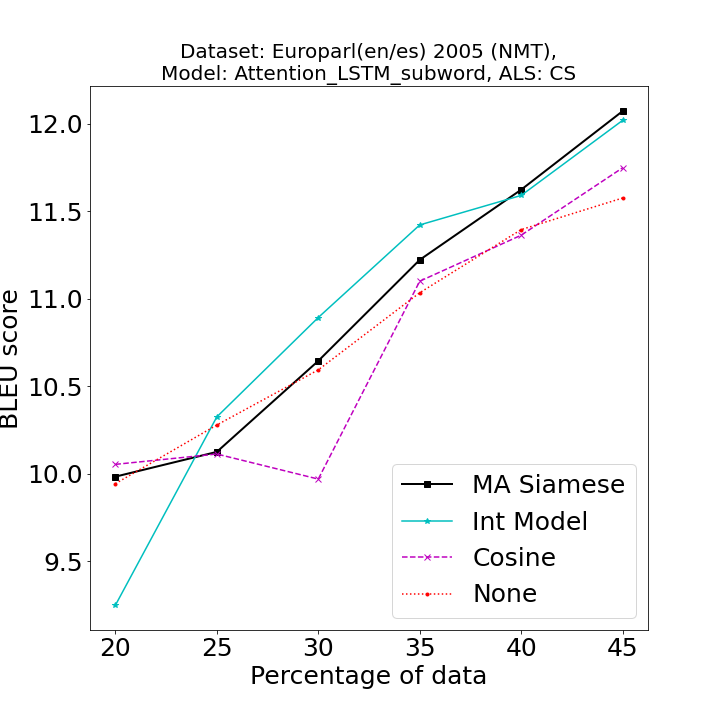}
		%\caption{Plot 11}
	\end{subfigure}%
	~
	\begin{subfigure}[t]{0.3\textwidth}
		\centering
		\includegraphics[width=\linewidth]{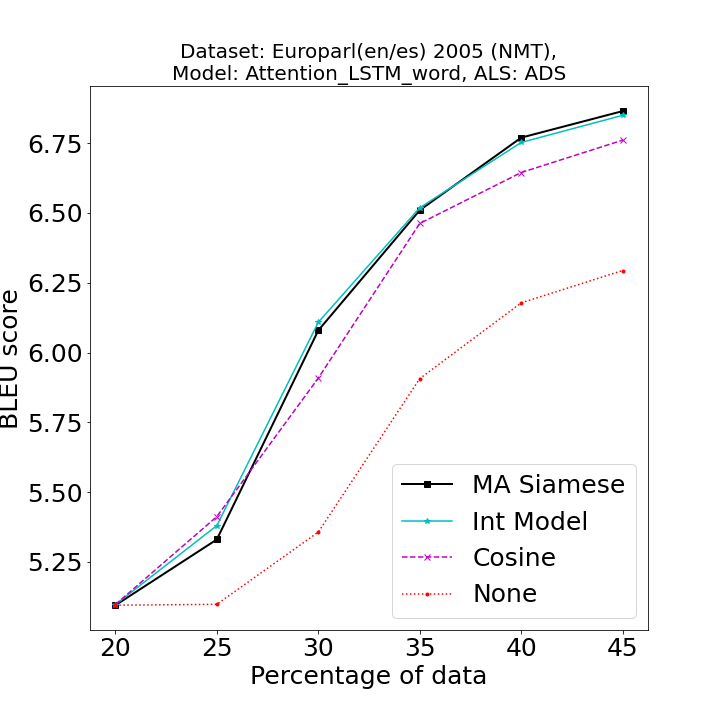}
		%\caption{Plot 12}
	\end{subfigure}
    \caption{[Best viewed in color] Comparison of our approach (A$^\mathbf{2}$L) with baseline approaches on different tasks using different active learning strategies. $1^{st}$ row: POS (error bound at convergence: $\pm 0.05$), $2^{nd}$ row: NER ($\pm 0.08$), $3^{rd}$ row: SEMTR ($\pm 0.09$), $4^{th}$ row: NMT ($\pm 0.04$). In the first three rows, from left to right, the three columns represent BALD, Entropy, and Margin AL strategies. $4^{th}$ row represents AL strategies for NMT, from left to right (LC: Least Confidence, CS: Coverage Sampling, ADS: Attention Distraction Sampling) :  Legend Description \{100\% data: full data performance, A$\mathbf{^2}$L (MA Siamese) : Model Aware Siamese, A$\mathbf{^2}$L (Int Model) : Integrated Clustering Model, Cosine : Cosine similarity, None : Active learning strategy without clustering step, Random : Random split (no active learning applied)\}. See Section \ref{section:baselines} for more details about the baselines. All the results were obtained by averaging over 5 random splits. These plots have been magnified to highlight the regions of interest. \what{For original plots, refer to Fig~\ref{original_graphs} in the Appendix.}}
	\label{magnified_graphs}
\end{figure*}

%==========================================%
\begin{figure*}
	\centering
	\begin{subfigure}[t]{0.3\textwidth}
		\centering
		\includegraphics[width=\linewidth]{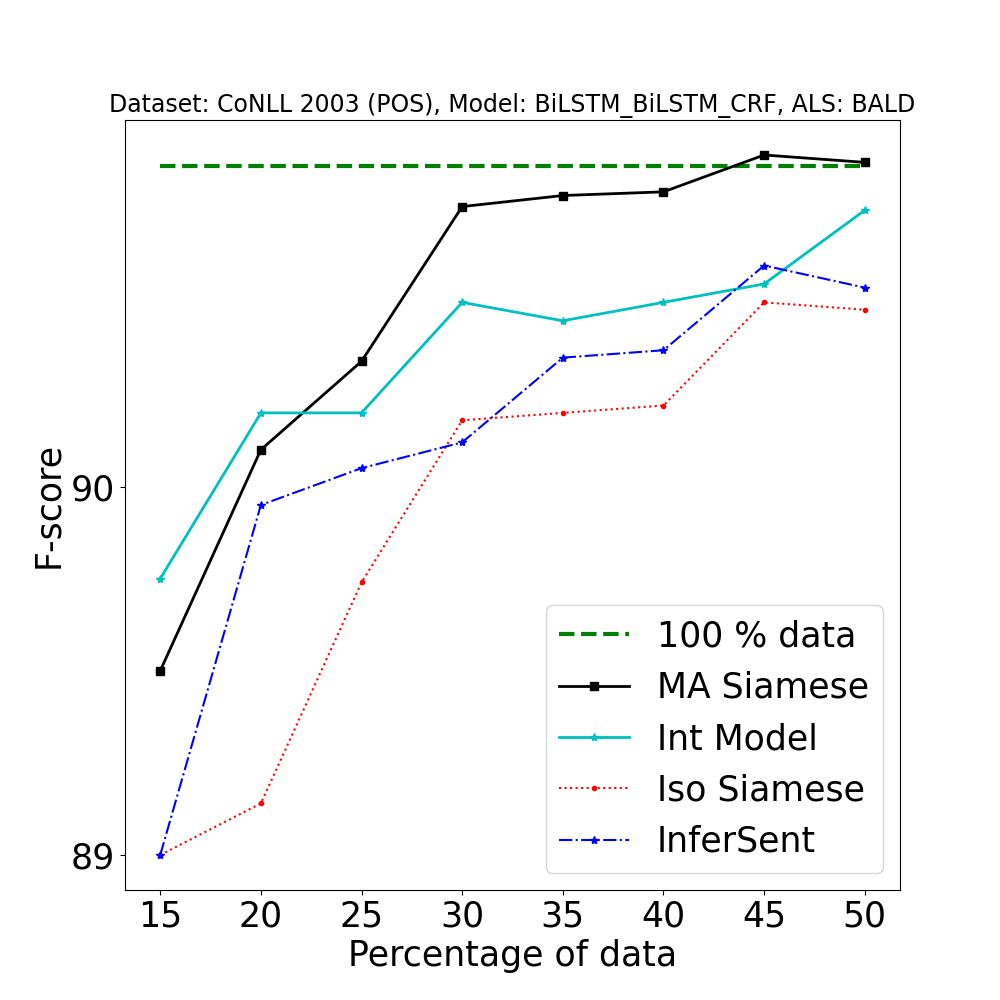}
	\end{subfigure}%
	~ 
	\begin{subfigure}[t]{0.3\textwidth}
		\centering
		\includegraphics[width=\linewidth]{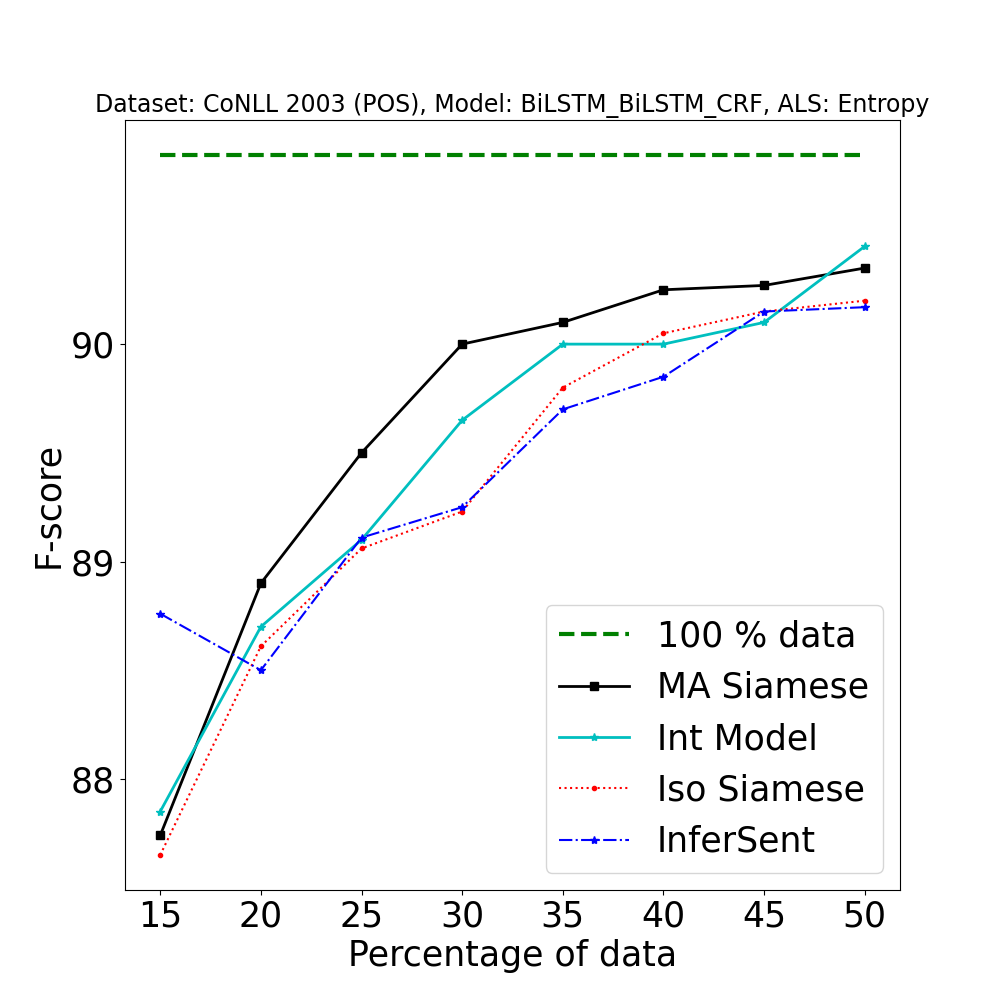}
	\end{subfigure}%
	~
	\begin{subfigure}[t]{0.3\textwidth}
		\centering
		\includegraphics[width=\linewidth]{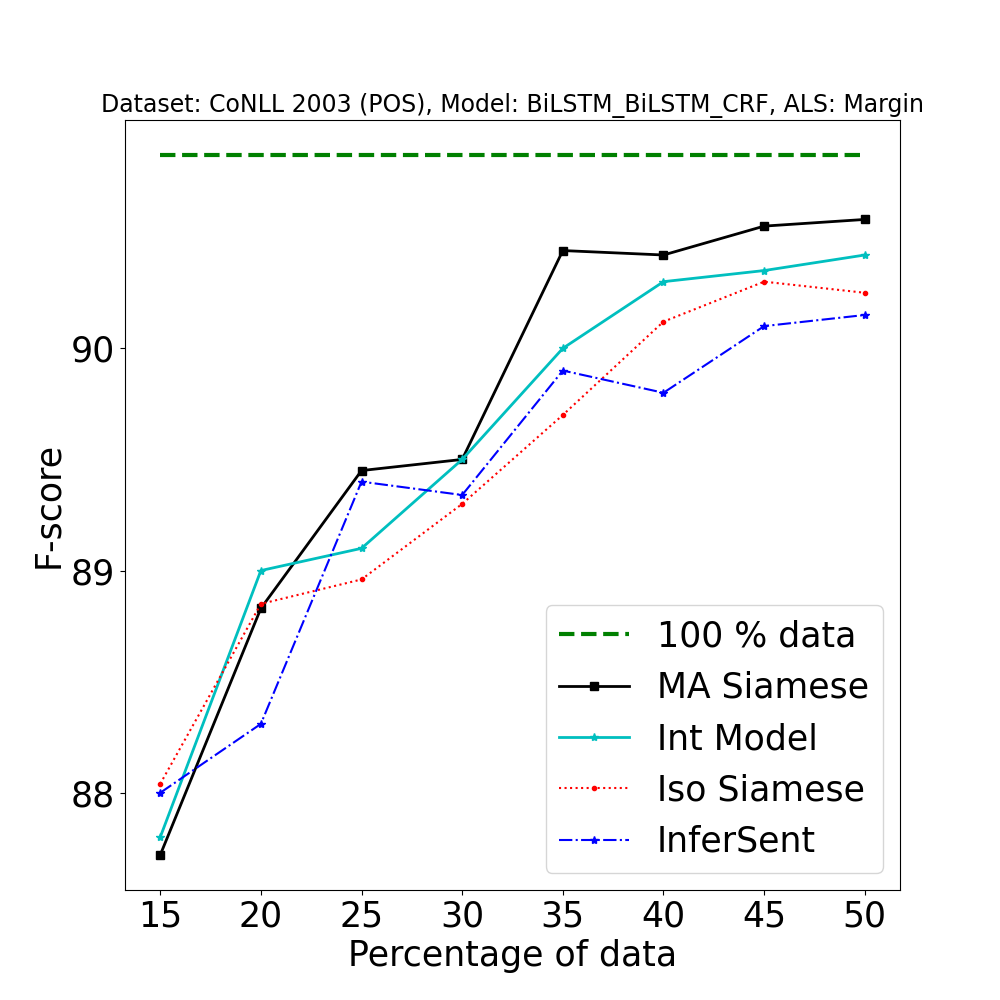}
	\
	\end{subfigure}%
	\caption{[Best viewed in color] Ablations studies on POS task using different active learning strategies. From left to right, the three columns represent BALD, Entropy and Margin based AL strategies. Legend Description \{\textbf{100\%} data : full data performance, A$\mathbf{^2}$L (MA Siamese) : Model Aware Siamese, A$\mathbf{^2}$L (Int Model) : Integrated Clustering Model, Iso Siamese : Model isolated Siamese, InferSent : Cosine similarity based on InferSent encodings\}. See Figure~\ref{magnified_graphs_ablations_all} in Appendix for experiments on other tasks. All the results were obtained by averaging over 5 splits. }
	\label{magnified_graphs_ablation}
\end{figure*}

\subsection{Baselines}
\label{section:baselines}

We claim that A$^2$L mitigates the redundancies in the existing AL strategies by working in conjunction with them. We validate our claims by comparing our approaches with three baselines that highlight the importance of various components.
\vspace{-0.2cm}
  \paragraph{\textbf{Cosine}:} Clustering is done based on cosine similarity between last output encodings (corresponding to sentence length) from encoder in $\mathcal{M}$. Although this similarity computation is model-aware, it is simplistic and shows the benefit of using a more expressive similarity measure.
\vspace{-0.2cm}  
  \paragraph{\textbf{None}:} In this baseline, we use the AL strategy without applying Active$^2$ learning to remove redundant examples. This validates our claim about redundancy in examples chosen by AL strategies. 
\vspace{-0.2cm}
  \paragraph{\textbf{Random}:} No active learning is used, and random examples are selected at each time.
% \end{itemize}

% ========================================================
%============================================%

\subsection{Ablation Studies}
\label{subsection:ablation study}
We perform ablation studies to demonstrate the utility of model-awareness using these baselines:
\vspace{-0.2cm}
      \paragraph{\textbf{Infersent}:} Clustering is done based on cosine similarity between sentence embeddings \cite{Chen:2015:SAL:2868301.2868619} obtained from a pre-trained InferSent model \cite{ConneauEtAl:2017:Supervised}. This similarity computation is not model-aware and shows the utility of model-aware similarity computation.
 \vspace{-0.2cm}    
      \paragraph{\textbf{Iso Siamese}:} To show that the Siamese network alone is not sufficient and model-awareness is needed, in this baseline, we train the Siamese network by directly using GloVe embeddings of the words as input rather than using output from the model $\mathcal{M}$'s encoder. This similarity, which is not model-aware, is then used for clustering.
 
\begin{figure}[!h]
    \centering  
    \includegraphics[width=\linewidth]{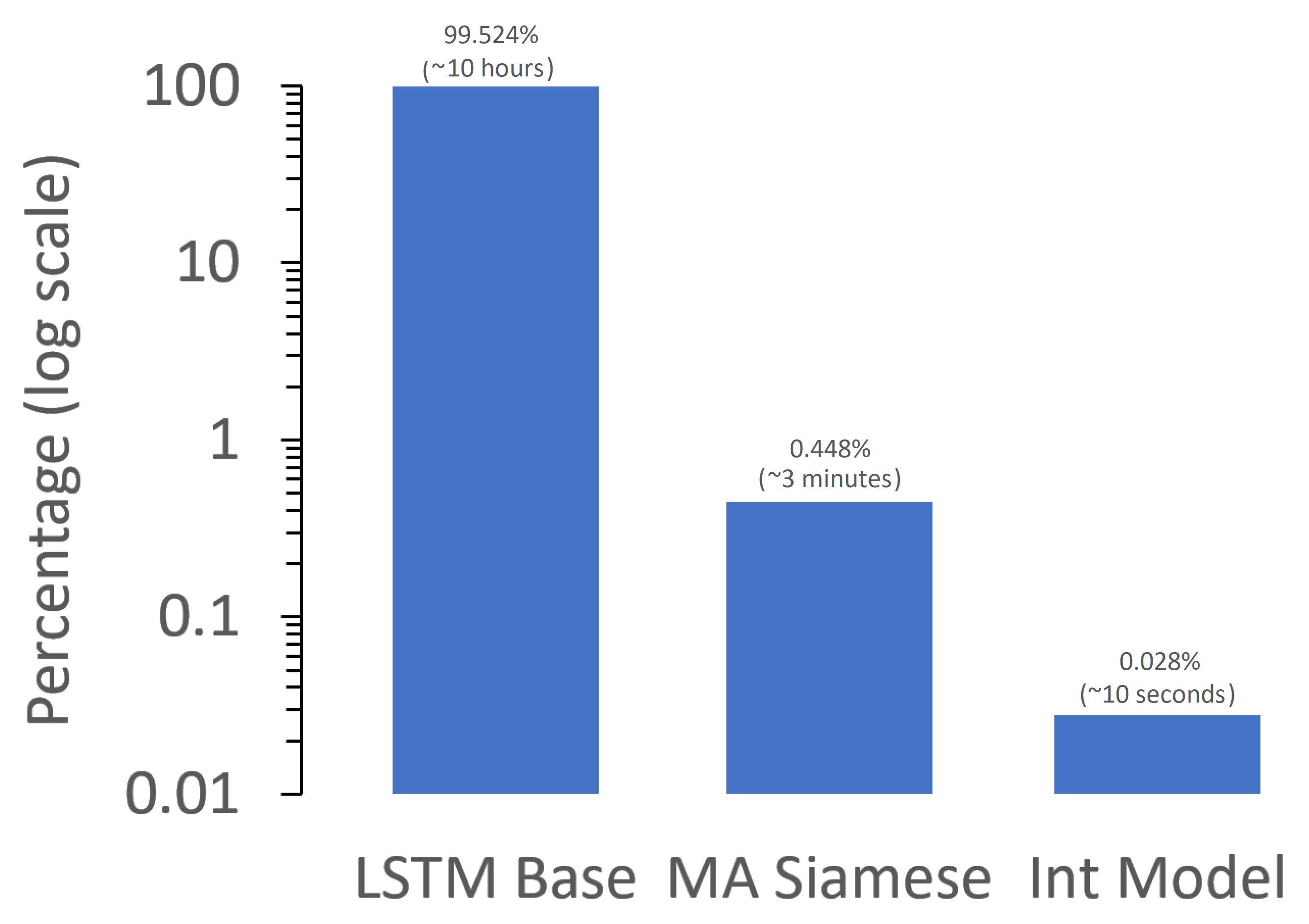}
    \caption{
    % Comparison of training time taken for one epoch in NMT full training by the various models at different stages of the pipeline, namely
    Comparison between various components of our approach in terms of the time required by them per training epoch for NMT.
    LSTM (Base) encoder-decoder translation model ($\approx 10$ hours), Model Aware (MA) Siamese ($\approx 3$ minutes) and Integrated Clustering (Int) Model ($\approx 10$ seconds). It can be observed that A$^2$L adds a negligible overhead to the overall training time ($\approx 0.45\%$ of the time taken by the base model).}
    \label{bar_plot_training_comparison}
\vspace{-0.5cm}
\end{figure}

\section{Results}
\label{section:results}
Figure~\ref{magnified_graphs} compares the performance of our methods with baselines. It shows the test-set metric on the $y$-axis against the percentage of training data used on the $x$-axis for all tasks. See Figures \ref{magnified_graphs_ablations_all} and \ref{original_graphs} in the Appendix for additional results.

%=====================================================%         

\begin{figure*}[t]
    \centering
    \includegraphics[width=\textwidth]{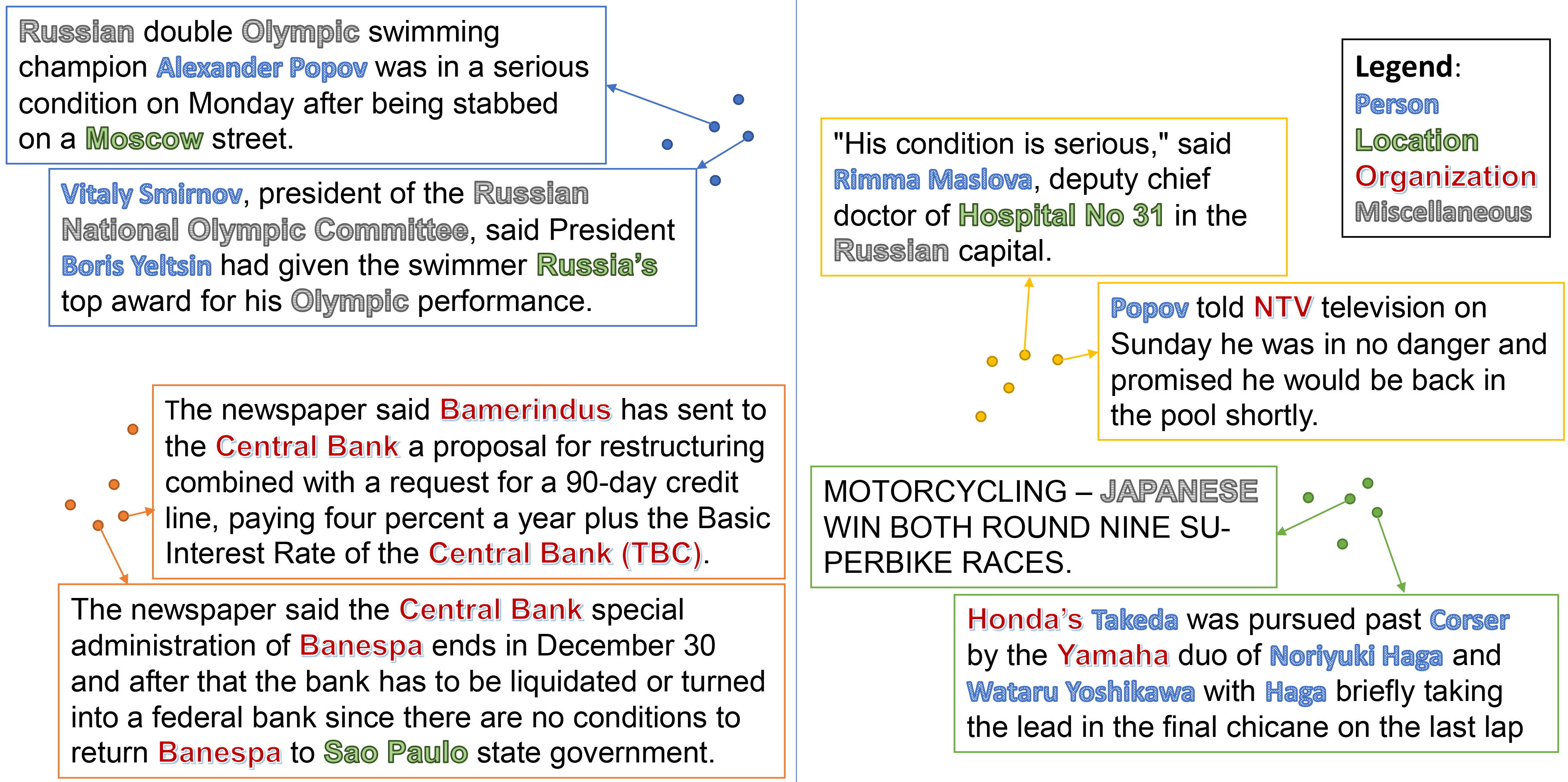}
    \caption{[Best viewed in color] Qualitative case study to convey the notion of redundancy and the model aware similarity. We compare the examples grouped in the same cluster using (i) MA similarity scores [left] and (ii) Cosine similarity scores on the InferSent embedding (Infersent baseline described in Section \ref{section:baselines}) [Right]. As expected, when cosine similarity is used, sentences that have roughly similar content have been assigned to the same cluster. However, when model aware similarity is used, in addition to having similar content, the sentences also have a similar tagging structure. It is sensible to eliminate sentences having similar tagging structures, as they are redundant. [Dataset: CoNLL 2003 dataset, AL strategy: BALD, Data usage: $10\%$ of the total data]}
    \label{fig:qualitative_examples_ablation}
\vspace{-0.4cm}
\end{figure*}

\begin{enumerate}[leftmargin=*,noitemsep]
 
     \item As shown in Figure~\ref{magnified_graphs}, our approach consistently outperforms all baselines on chosen tasks. Note that one should observe how fast the performance increases with the addition of training data (and not just the final performance) as we are trying to evaluate the effect of adding new examples. Our ablation studies in Figure~\ref{magnified_graphs_ablation} show the utility of using model-aware similarity. 
     
     \item In sequence tagging, we match the performance obtained by training on the full dataset using only a smaller fraction of the data ($\mathbf{3-25\%}$ less data as compared to state-of-art AL strategies) (Table~\ref{tab_results}). On a large dataset in NMT task (Europarl), A$^\mathbf{2}$L takes $\approx 4300$ sentences fewer than the Least Confidence AL strategy to reach a Bleu score of $12$.
     
     \item While comparing different AL strategies is not our motive, Figure~\ref{magnified_graphs} also demonstrates that one can achieve performance comparable to a complex AL strategy like BALD, using simple AL strategies like margin and entropy, by using the proposed A$^\mathbf{2}$L framework. 
     
     \item Additionally, from Figure~\ref{bar_plots_comparison}, it can be observed that for one step of data selection: (i) The proposed MA Siamese model adds minimal overhead to the overall AL pipeline since it takes less than 5 additional seconds ($\approx \frac{1}{12}$ of the time taken for ALS); (ii) By approximating the clustering step, Integrated Clustering (Int) Model further reduces the overhead down to 2 seconds. However, owing to this approximation, MA Siamese is observed to perform slightly better than the Int Model (Fig~\ref{magnified_graphs_ablation}). A comparison of training time for various stages of the A$^2$L pipeline is provided in Figure~\ref{bar_plot_training_comparison}.
     
 \end{enumerate} 

We wish to state that our approach should be evaluated not in terms of the gain in the F1 score but in terms of the reduction in data required to achieve the same (3-25 \% on multiple datasets). More importantly, this improvement comes at a negligible computation overhead cost. The reported improvements are not relative with respect to any baseline but represent an absolute value and are very significant in the context of similar performance improvements reported in the literature. In Figure \ref{fig:qualitative_examples_ablation}, we provide a qualitative case study that demonstrates the problem of redundancy.

\begin{table}
    \centering
    \begin{tabular}{ >{\centering\arraybackslash}m{1.9cm} >{\centering\arraybackslash}m{2.2cm} >{\centering\arraybackslash}m{2.3cm}}
        \toprule
         No. of examples selected by AL Strategy & Spectral (examples processed per second) & Integrated (examples processed per second)  \\
         \midrule
         5000 & 491.64 & 2702.70 \\
         10000 & 248.08 & 2557.54 \\
         15000 & 163.10 & 2508.36 \\
         20000 & 122.89 & 2493.76 \\
         \bottomrule
    \end{tabular}
    \caption{Number of samples processed per second by the Spectral Clustering (MA Siamese) compared to the Integrated Clustering (Int Model) methods.}
    \label{tab:speed_advantages_of_intcluster}
    \vspace{-0.5cm}
\end{table}

\section{Conclusion}
\label{section:conclusion}
This paper shows that one can further reduce the data requirements of Active Learning strategies by proposing a new method, A$^2$L, which uses a model-aware-similarity computation. We empirically demonstrated that our proposed approaches consistently perform well across many tasks and AL strategies. We compared the performance of our approach with strong baselines to ensure that the role of each component is properly understood.

\section*{Acknowledgement}
This work was funded by British Telecom India Research Center project on Advanced Chatbot.

\bibliography{biblio}
\bibliographystyle{acl_natbib}

%%%%%%%%%%%%%%%%%%%%%%%%%%%%%%%%%%%%%%%%%%%%%%%%%%%%%%%%%%%%%%%%%%%%%%%%%%%%%%%%

\newpage
% \textcolor{white}{.}
% \newpage
\twocolumn[
\begin{center}
\textbf{\Large Active$^2$ Learning: Actively reducing redundancies in Active Learning methods for Sequence Tagging and Machine Translation: \\Appendix }
\end{center}
\hfill \break
\hfill \break
]

\appendix
\label{sec:appendix}

\section{Additional Remarks}
\label{appendix:additional remarks}
In this section, we make a number of additional remarks about the proposed approach.

\subsection{What is the significance of our work?}
\label{section:what_is_the_significance_of_our_work}
Obtaining labeled data is both time-consuming and costly. Active learning is employed to minimize the labeling effort. However, as we point out in Section~\ref{section:introduction}, existing techniques may select redundant examples for manual annotation. Due to this redundancy, there is a scope for improvement in the performance of active learning strategies, and our proposed approach fills this gap. Since we demonstrate that our method is compatible with many active learning strategies and deep learning models that are currently in use, it can be applied in a wide range of contexts and is likely to be useful for many sub-communities within the domain of natural language processing without adding significant complexity to the existing systems.

\subsection{How do we validate our claim regarding the sub-optimality of standard AL strategies due to redundancy?}
\label{section:validate_claim_regarding_sub_optimality}
The comparison of our approach with None baseline suggests that performance comparable to the state-of-art can be achieved by using fewer labels if one incorporates the second step, which eliminates allegedly redundant examples even when every other aspect of training is exactly the same (same model, AL strategy and dataset). Thus, we can say that the discarded examples were of no additional help for the model and hence were redundant. Avoiding annotation of such samples saves time and brings down both computational and annotation costs. This can especially be effective in, for instance, the medical domain where high expertise is required.

%%%%%%%%%%%%%%%%%%%%%%%%%%%%%%%%%%%%%%%%%%%%%%%%%%%%%%%%%%%%%%%%%%%%%%%%%%%%%%%%%%%%%%
\section{Hyper-parameters and other Implementation Details}

\begin{figure}[h!]
\centering  
\includegraphics[width=1.05\linewidth]{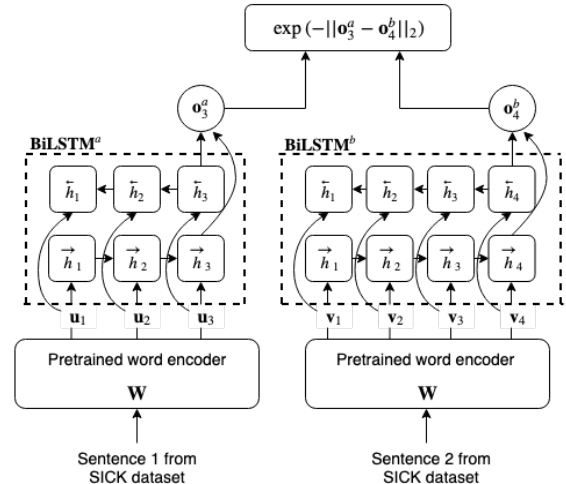}
\caption{Modeling similarity using the Siamese encoder (enclosed by dotted lines). A pair of sentences from SICK dataset is fed to the pretrained sequence tagging model. The output of the word encoder is then passed to the Siamese encoder. Last hidden state of the Siamese encoder, corresponding to the sequence length of the sentence, is used for assigning a similarity score to the pair.}
\label{appendix:fig_siamese}
\end{figure}

Similar hyper-parameter values work across all the tasks. Hence, the same values were used for all experiments and these values were determined using the validation set of the CoNLL 2003 dataset for NER task. We use two different sequence tagging architectures: CNN-BiLSTM-CRF model (CNN for character-level encoding and BiLSTM for word-level encoding) and a BiLSTM-BiLSTM-CRF model~\cite{N16-1030} (BiLSTM for both character-level and word-level encoding). The CNN-BiLSTM-CRF architecture is a light-weight variant of the model proposed in~\cite{SiddhantEtAl:2018:Deep}, having one layer in CNN encoder with two filters of sizes 2 and 3, followed by a max pool, as opposed to three layers in the original setup. This modification was found to improve the results. We use glove embeddings~\cite{pennington2014glove} for all datasets. We apply normal dropout in the character encoder instead of the use of recurrent dropout~\cite{NIPS2016_6241} in the word encoder of the model presented in~\cite{SiddhantEtAl:2018:Deep} owing to an improvement in performance. For numerical stability, we use log probabilities and, thus, the value for margin-based AL strategy's threshold is outside the interval $[0, 1]$. We use the spectral clustering \cite{NIPS2001_2092} algorithm to cluster the sentences chosen by the AL strategy. We chose two representative examples from each cluster.\\

\label{Table of hyperparameters}
\begin{tabular}{p{6cm} p{1cm}}
\hline
\textbf{Active Learning strategy} & \\
\hline
threshold (Margin) & 15\\
threshold (Entropy) & 40\\
threshold (BALD) & 0.2\\
dropout (BALD) & 0.5\\
number of forward passes (BALD) & 51\\
\hline
\textbf{Sequence tagging model} & \\
\hline
CNN filter sizes & [2,3]\\
training batch size & 12\\
% splits of train data & 50\\
number of train epochs & 16\\
dimension of character embedding & 100\\
learning rate (Adam) & 0.005\\
learning rate decay & 0.9\\

\hline
\textbf{Siamese encoder} & \\
\hline
training batch size & 48\\
number of train epochs & 41\\
train/dev split & 0.8\\
learning rate (Adam) & 1e-5\\
period (of retrain) & 10\\
\hline
\textbf{Clustering} & \\
\hline
Number of clusters & 20\\
\hline
\textbf{Training} & \\
\hline
Batch size & 12\\
\hline
\hline
\hline
\textbf{NMT model} & \\
\hline
% CNN filter sizes & [2,3]\\
training batch size & 128\\
% splits of train data & 50\\
number of train epochs & 20\\
dimension of (sub)word embedding & 256\\
learning rate (Adam) & 1e-3\\
% learning rate decay & 0.9\\

\hline
\textbf{Siamese encoder} & \\
\hline
training batch size & 1150\\
number of train epochs & 25\\
% train/dev split & 0.8\\
dimension of (sub)word embedding & 300\\
learning rate (Adam) & 1e-3\\
period (of retrain) & 3\\
\hline
\textbf{Clustering} & \\
\hline
Number of clusters & 50\\
\hline
\textbf{Training} & \\
\hline
Batch size & 128\\
\hline
\end{tabular}

\begin{figure*}
	\centering
	\begin{subfigure}[t]{0.3\textwidth}
		\centering
		\includegraphics[width=\linewidth]{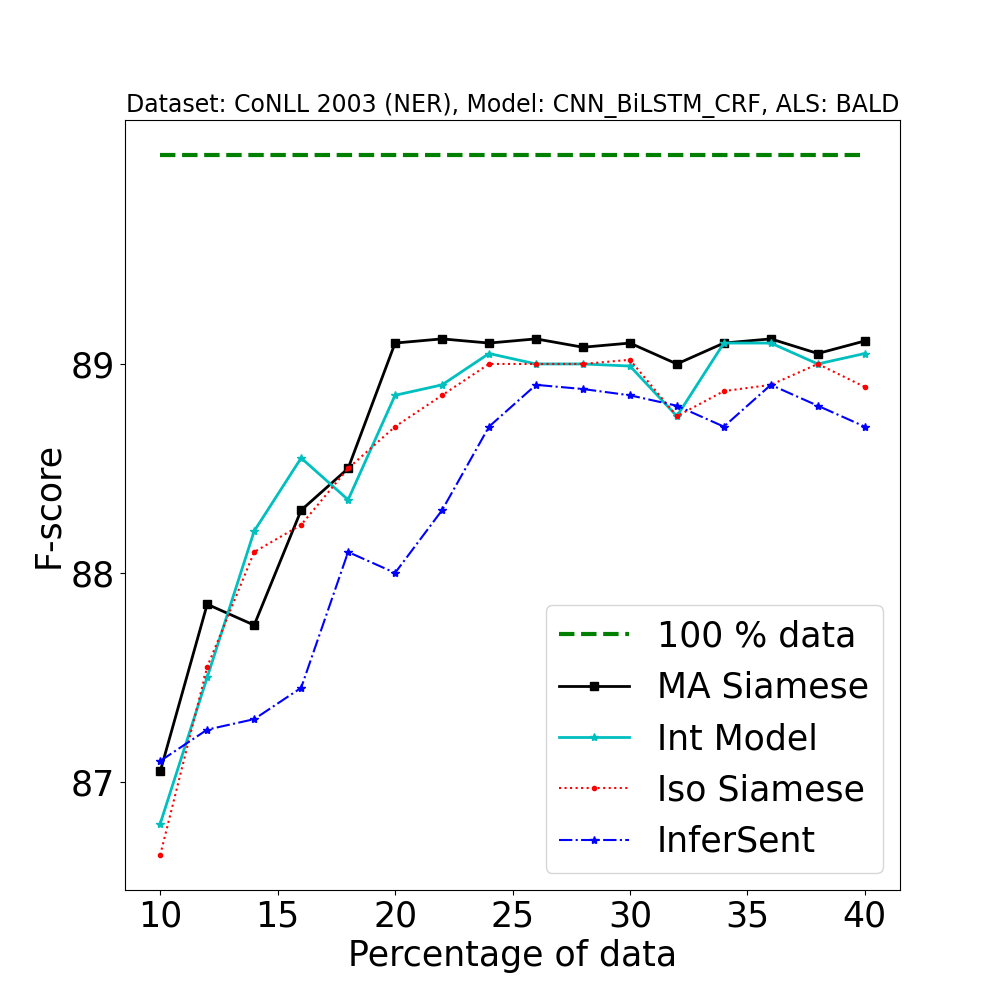}
		% \caption{Plot 4}
	\end{subfigure}%
	~ 
	\begin{subfigure}[t]{0.3\textwidth}
		\centering
		\includegraphics[width=\linewidth]{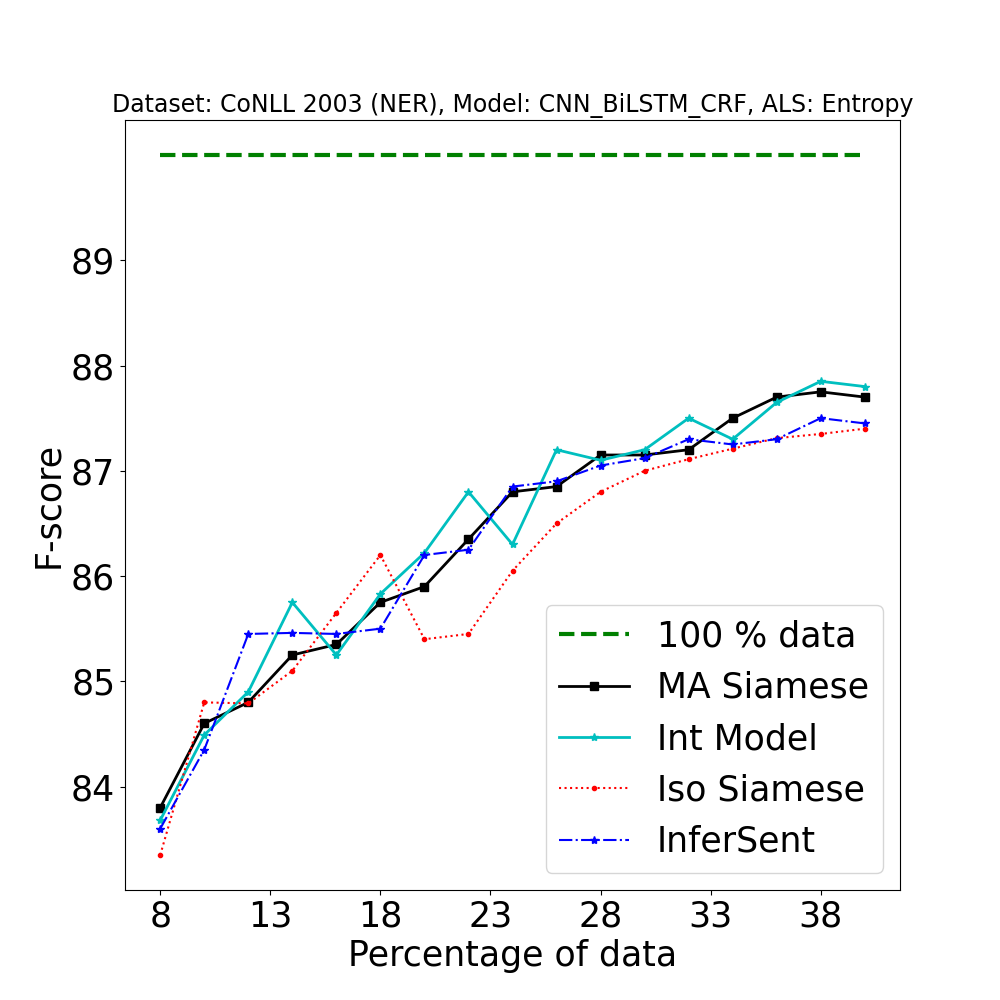}
		%\caption{Plot 5}
	\end{subfigure}%
	~
	\begin{subfigure}[t]{0.3\textwidth}
		\centering
		\includegraphics[width=\linewidth]{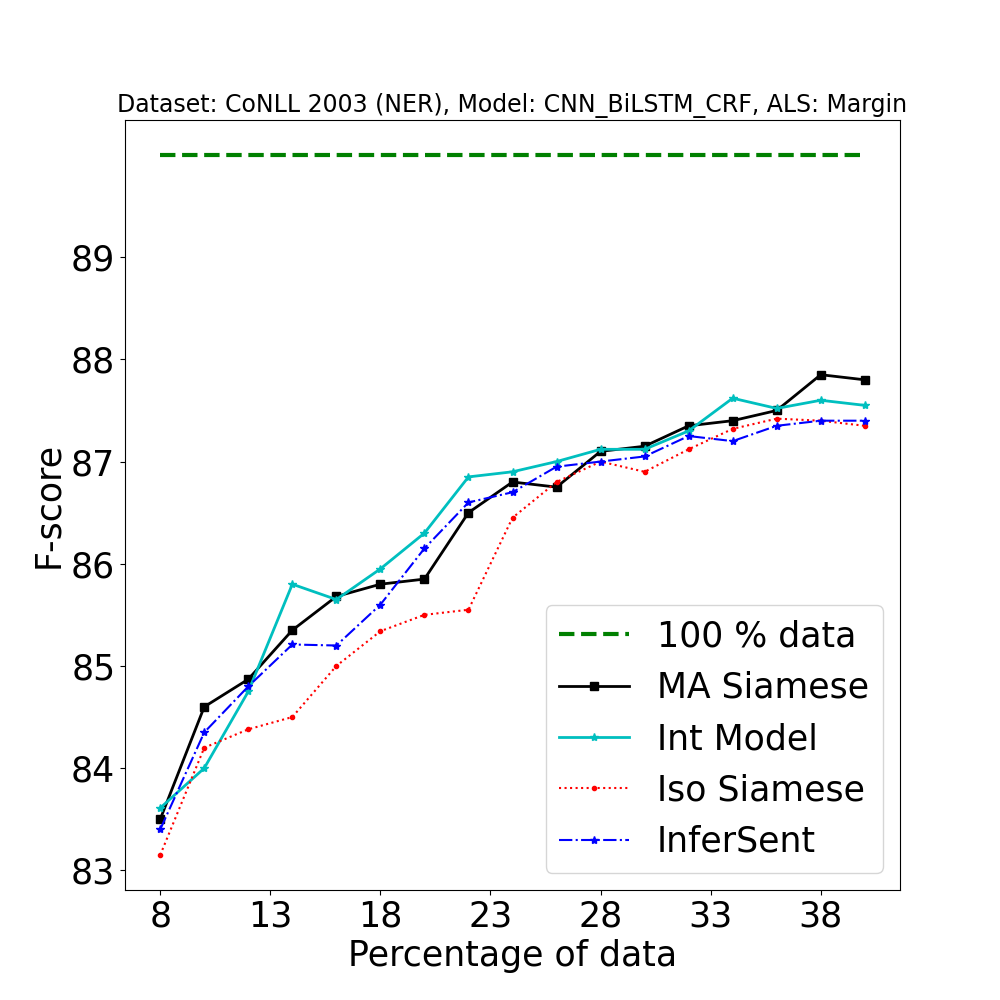}
		%\caption{Plot 6}
	\end{subfigure}%
	\\
	\begin{subfigure}[t]{0.3\textwidth}
		\centering
		\includegraphics[width=\linewidth]{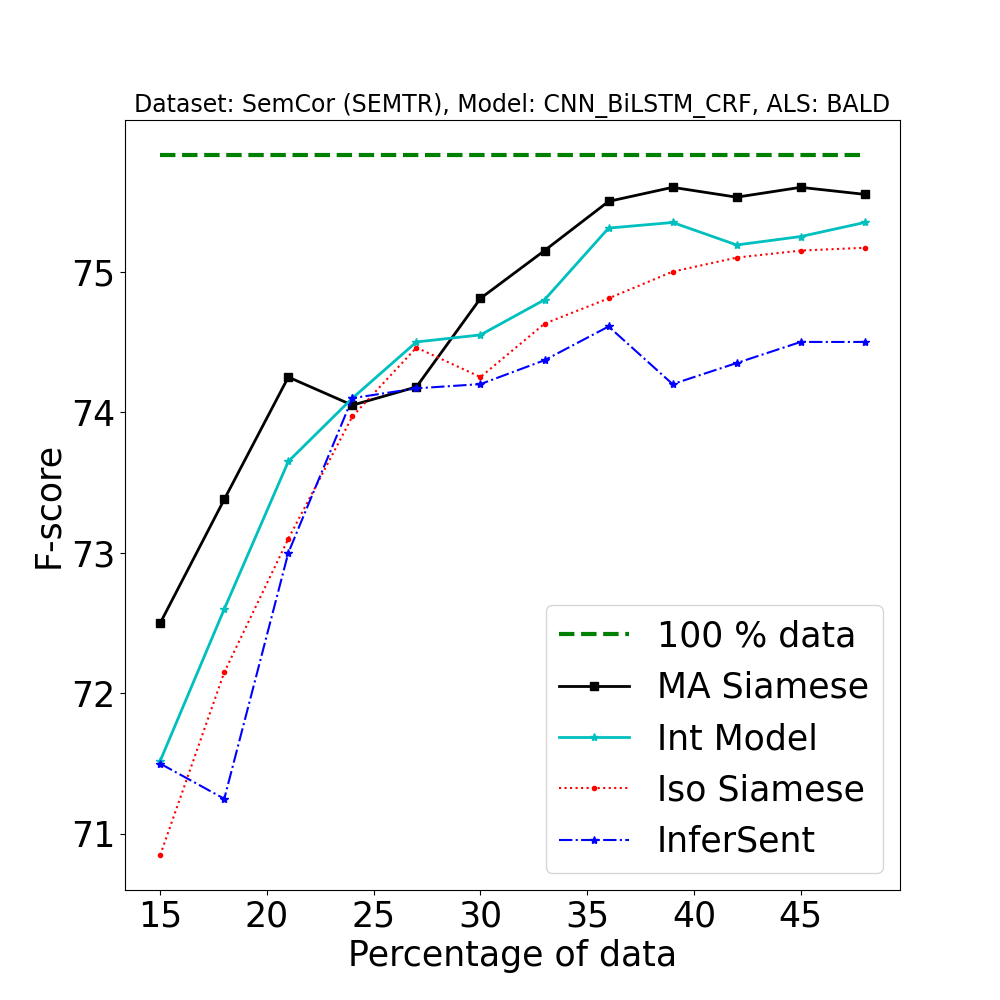}
		%\caption{Plot 1}
	\end{subfigure}%
	~ 
	\begin{subfigure}[t]{0.3\textwidth}
		\centering
		\includegraphics[width=\linewidth]{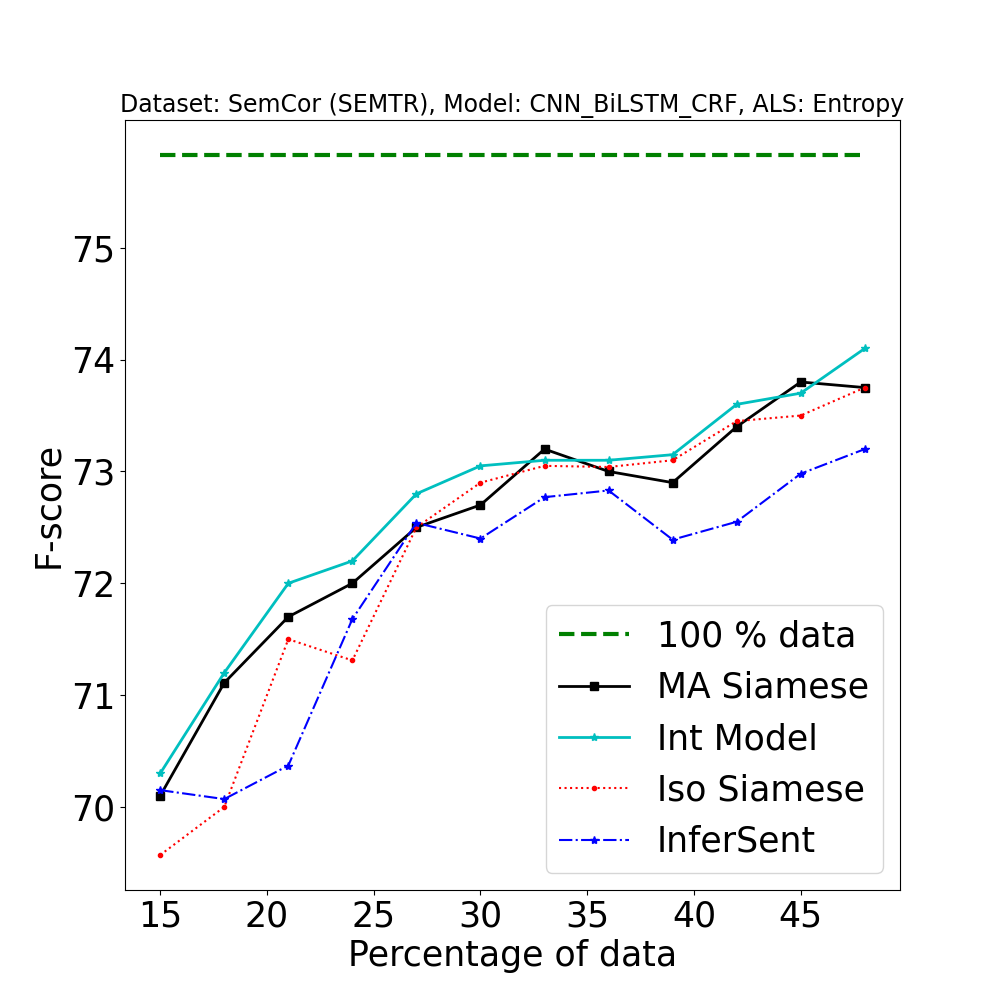}
		%\caption{Plot 2}
	\end{subfigure}%
	~
	\begin{subfigure}[t]{0.3\textwidth}
		\centering
		\includegraphics[width=\linewidth]{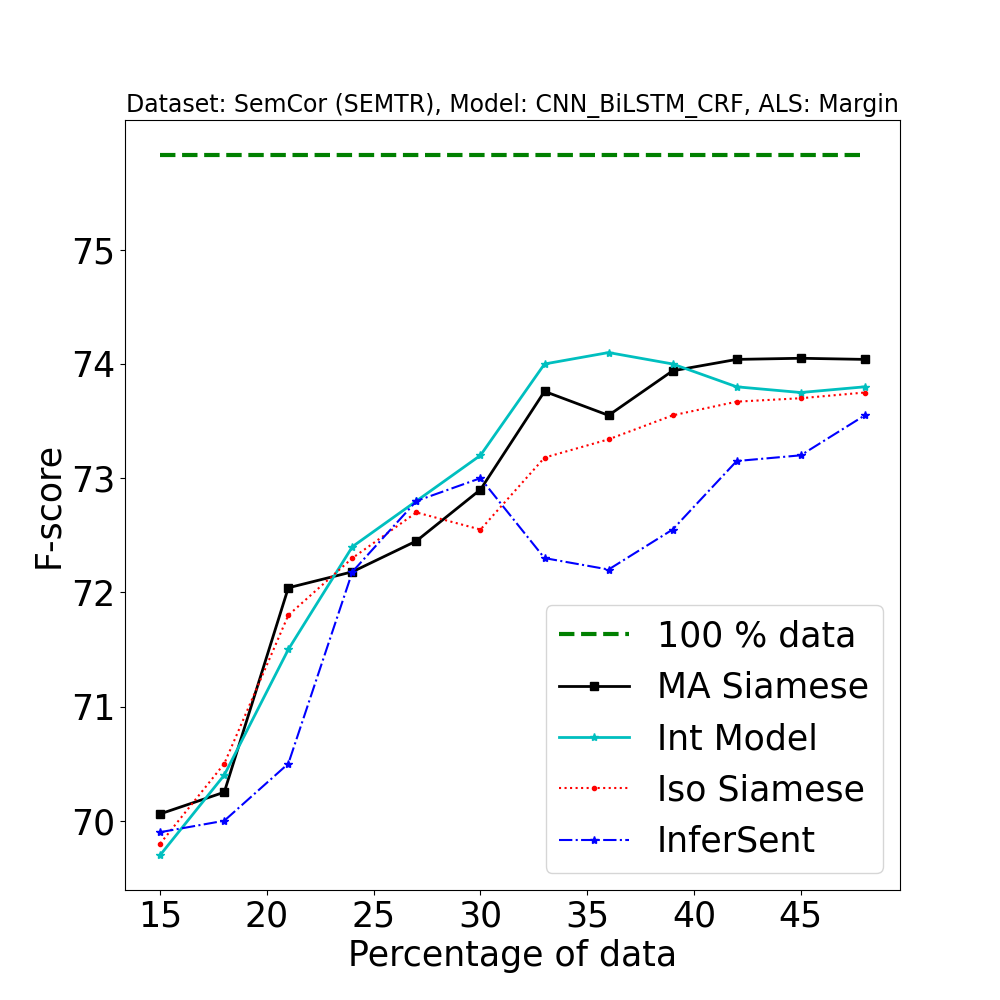}
		%\caption{Plot 3}
	\end{subfigure}%
	\\
	\begin{subfigure}[t]{0.3\textwidth}
		\centering
		\includegraphics[width=\linewidth]{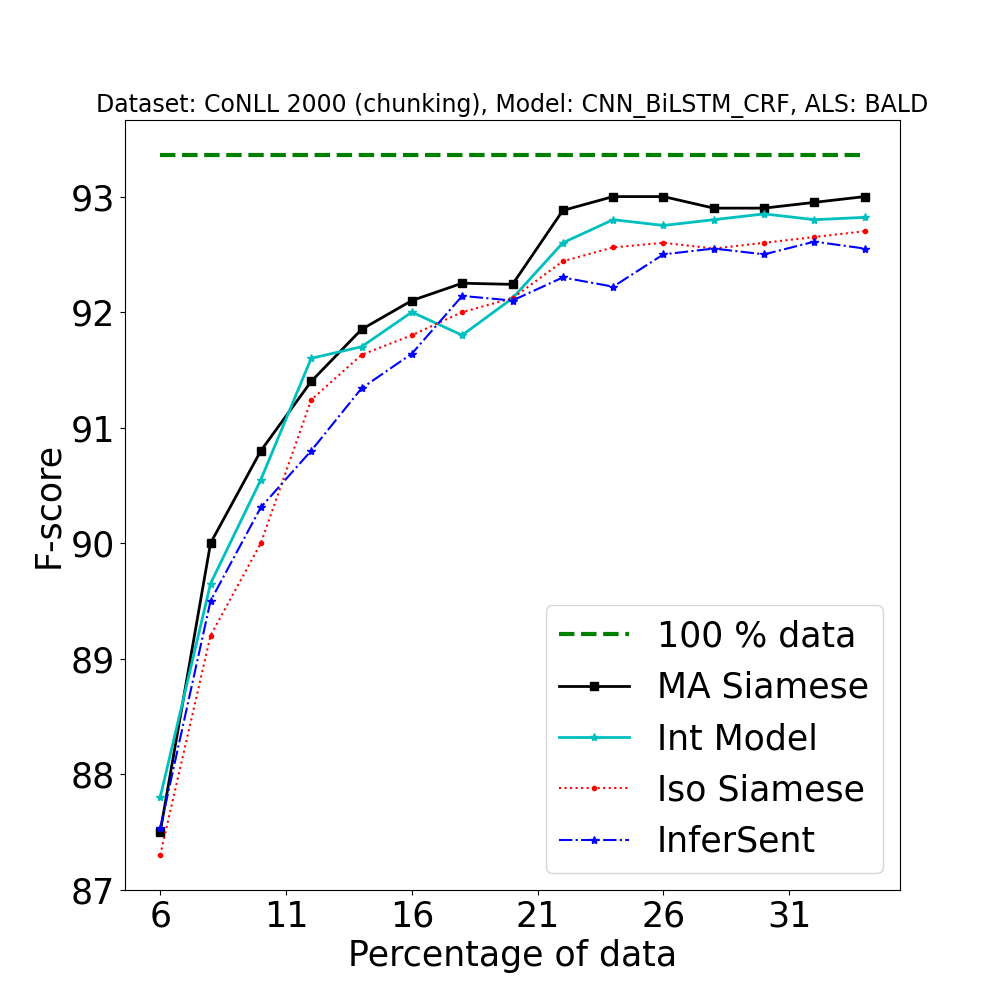}
		%\caption{Plot 10}
	\end{subfigure}%
	~ 
	\begin{subfigure}[t]{0.3\textwidth}
		\centering
		\includegraphics[width=\linewidth]{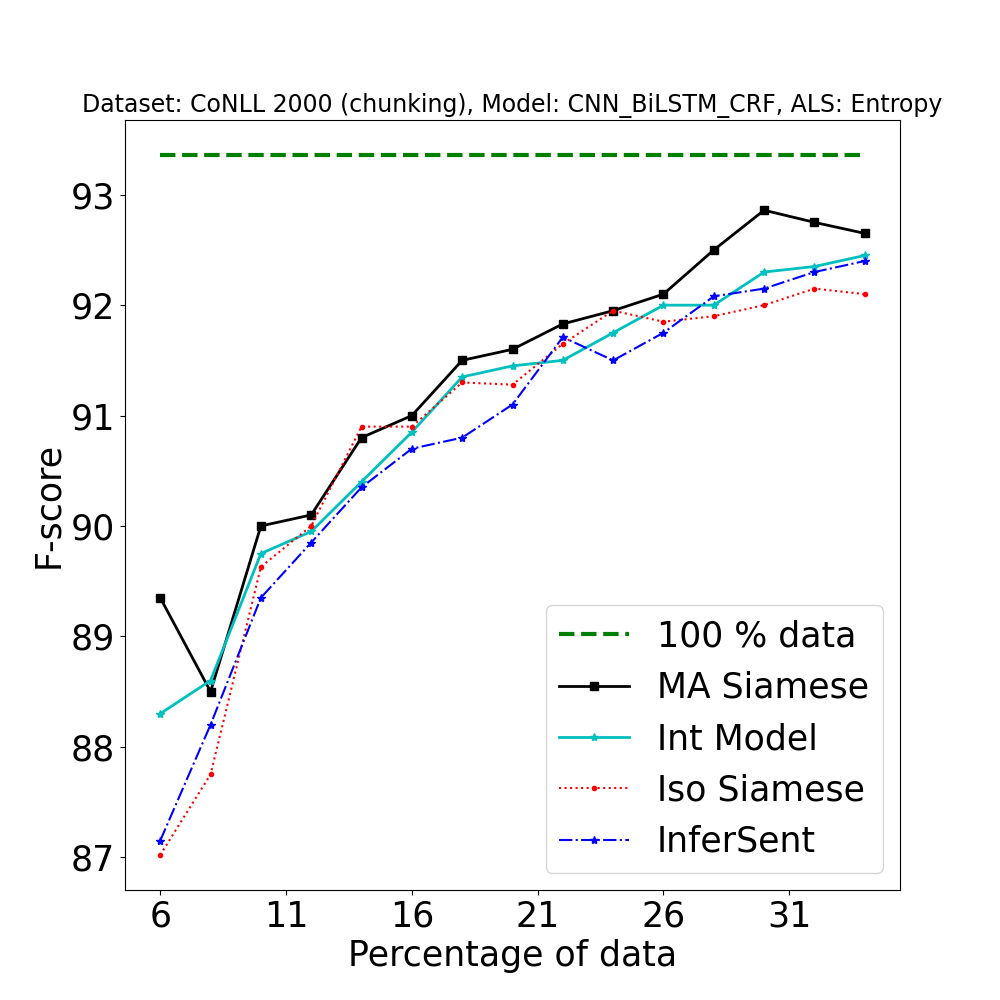}
		%\caption{Plot 11}
	\end{subfigure}%
	~
	\begin{subfigure}[t]{0.3\textwidth}
		\centering
		\includegraphics[width=\linewidth]{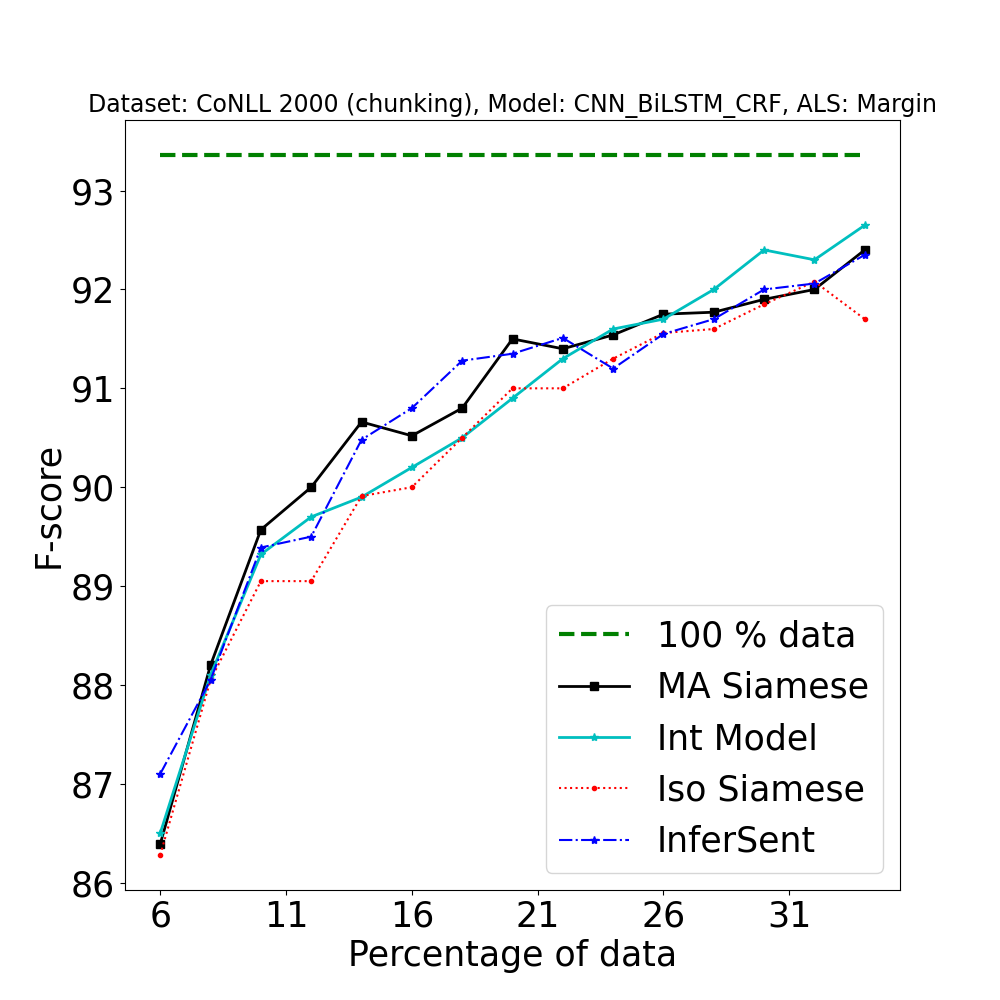}
		%\caption{Plot 12}
	\end{subfigure}
	\\
	\begin{subfigure}[t]{0.3\textwidth}
		\centering
		\includegraphics[width=\linewidth]{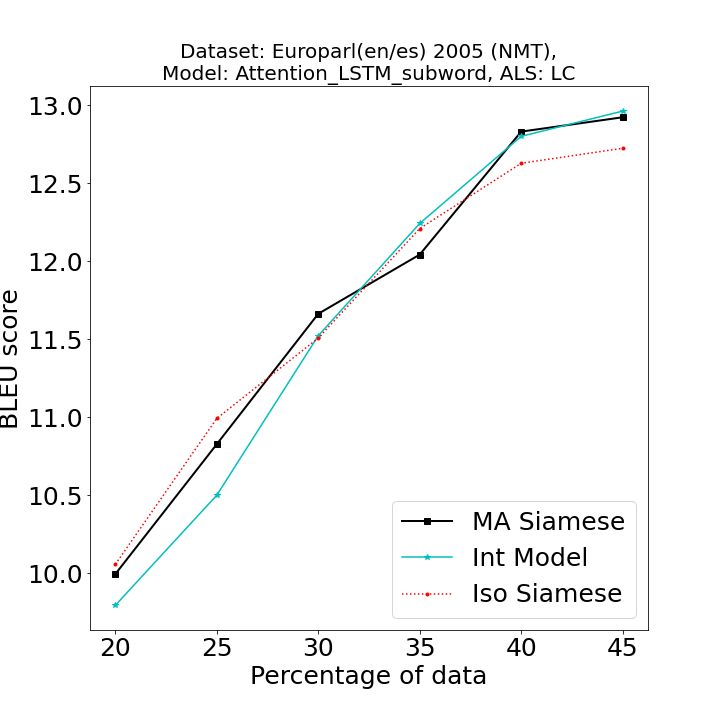}
		% \caption{Plot 4}
	\end{subfigure}%
	~ 
	\begin{subfigure}[t]{0.3\textwidth}
		\centering
		\includegraphics[width=\linewidth]{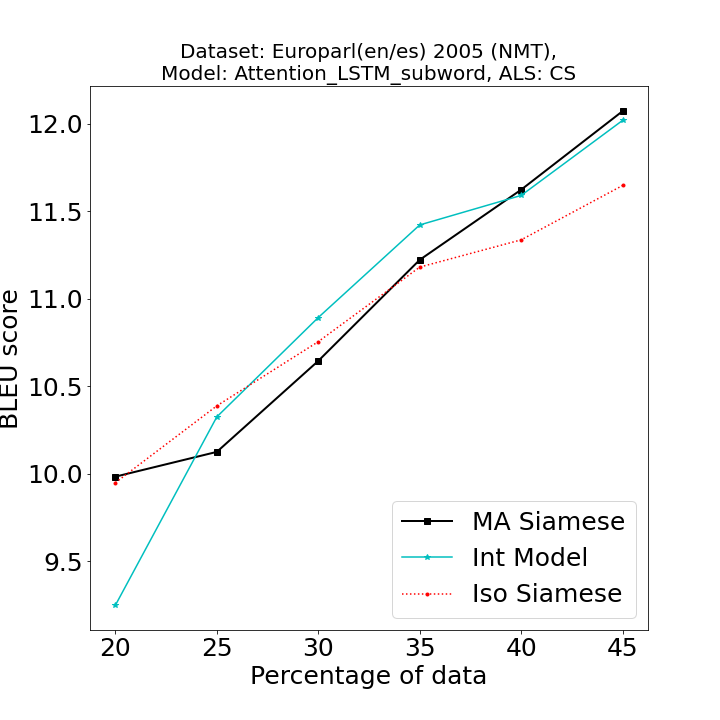}
		%\caption{Plot 5}
	\end{subfigure}%
	~
	\begin{subfigure}[t]{0.3\textwidth}
		\centering
		\includegraphics[width=\linewidth]{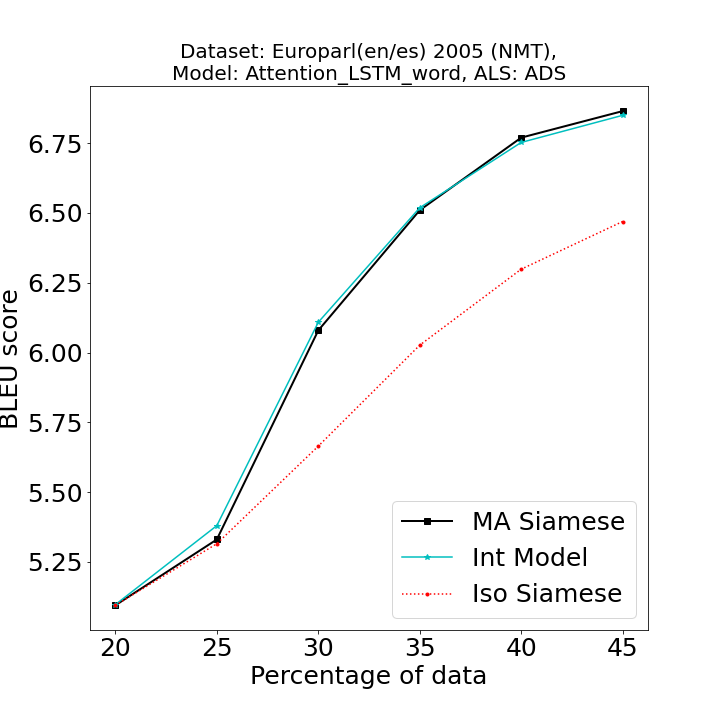}
		%\caption{Plot 6}
	\end{subfigure}%
	\caption{[Best viewed in color] Ablations studies on different tasks using different active learning strategies. $1^{st}$ row: NER, $2^{nd}$ row: SEMTR, $3^{rd}$ row: CHUNK, $4^{th}$ row: NMT. In first three rows, from left to right, the three columns represent BALD, Entropy and Margin AL strategies. $4^{th}$ row represents AL strategies for NMT, from left to right (LC: Least Confidence, CS: Coverage Sampling, ADS: Attention Distraction Sampling). Legend Description \{100\% data : full data performance, A$\mathbf{^2}$L (MA Siamese) : Model Aware Siamese, A$\mathbf{^2}$L (Int Model) : Integrated Clustering Model, Iso Siamese : Model isolated Siamese, InferSent : Cosine similarity based on InferSent encodings\}. See Section \ref{subsection:ablation study} for more details. All results were obtained by averaging over 5 random splits.}
	\label{magnified_graphs_ablations_all}
\end{figure*}
%%%%%%%%%%%%%%%%%%%%%%%%%%%%%%%%%%%%%%%%%%%%%%%%%%%%%%%%%%%%%

\begin{figure*}
	\centering
	\begin{subfigure}[t]{0.3\textwidth}
		\centering
		\includegraphics[width=\linewidth]{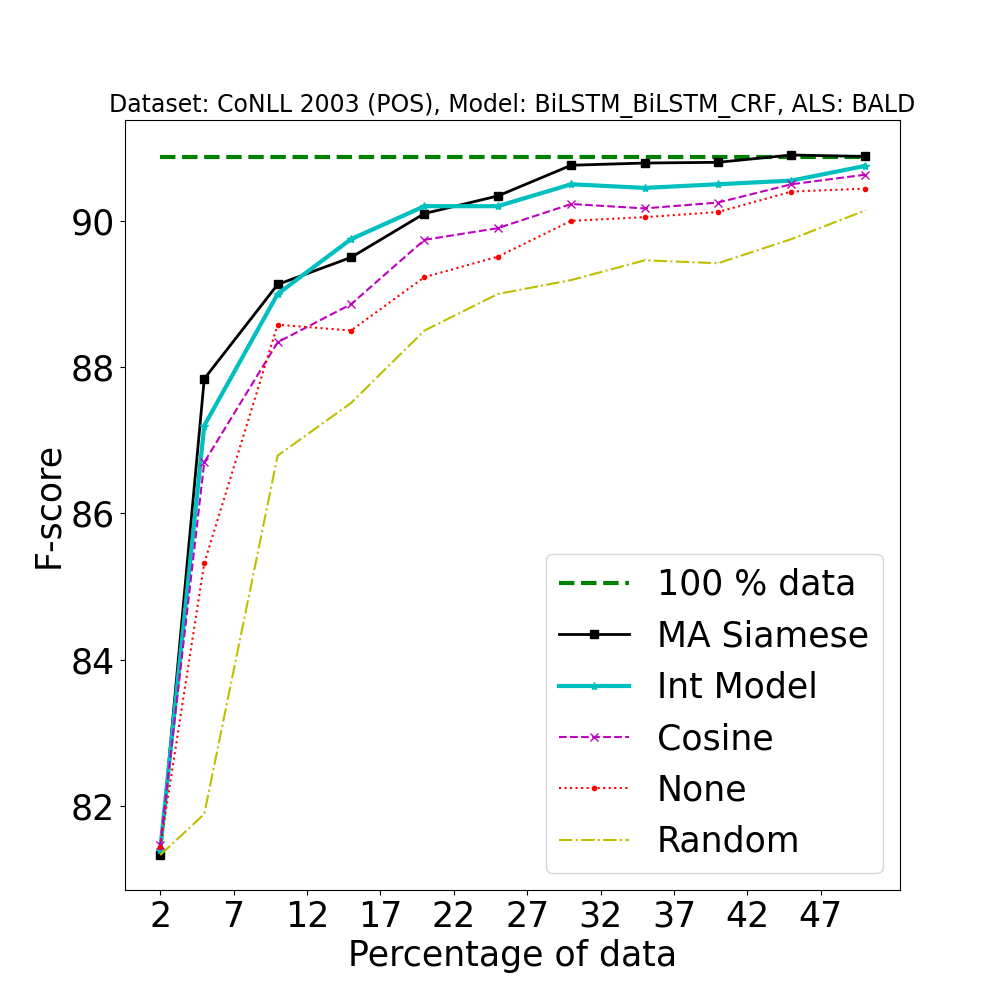}
		% \caption{Plot 4}
	\end{subfigure}%
	~ 
	\begin{subfigure}[t]{0.3\textwidth}
		\centering
		\includegraphics[width=\linewidth]{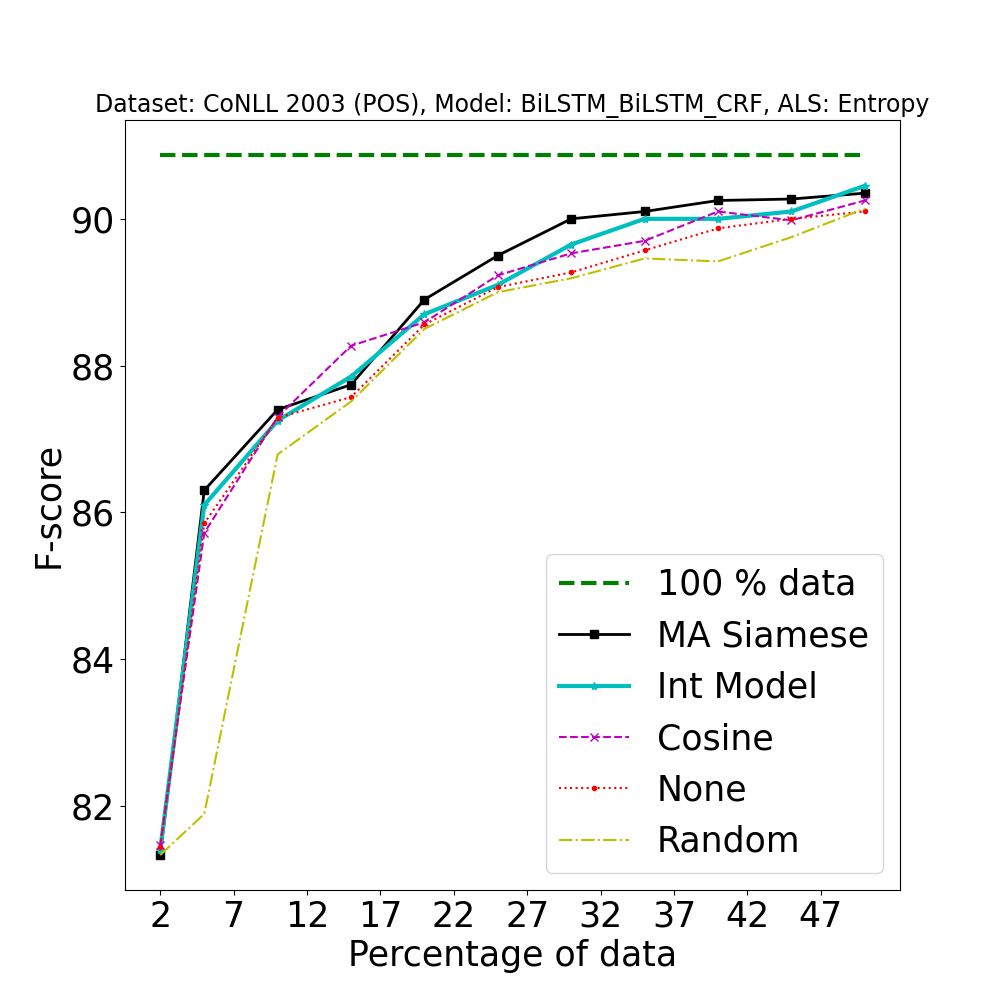}
		%\caption{Plot 5}
	\end{subfigure}%
	~
	\begin{subfigure}[t]{0.3\textwidth}
		\centering
		\includegraphics[width=\linewidth]{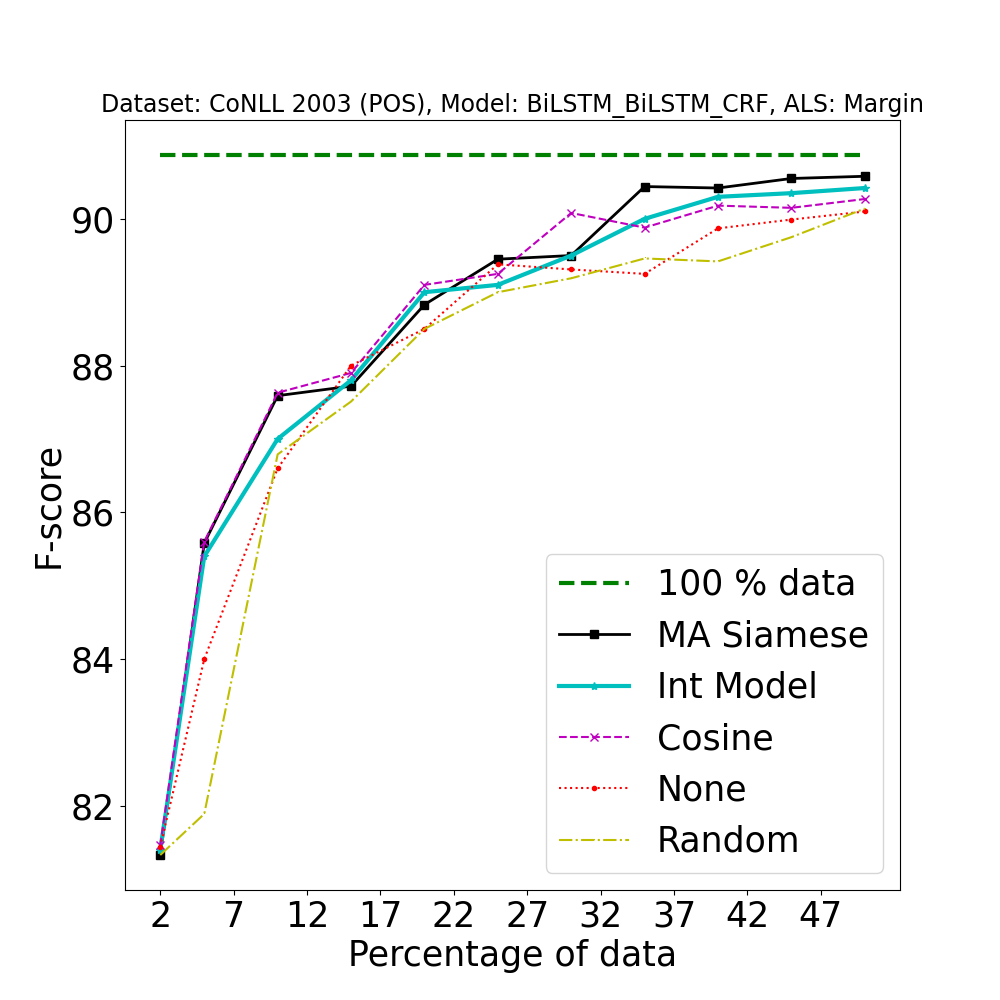}
		%\caption{Plot 6}
	\end{subfigure}%
	\\
	\begin{subfigure}[t]{0.3\textwidth}
		\centering
		\includegraphics[width=\linewidth]{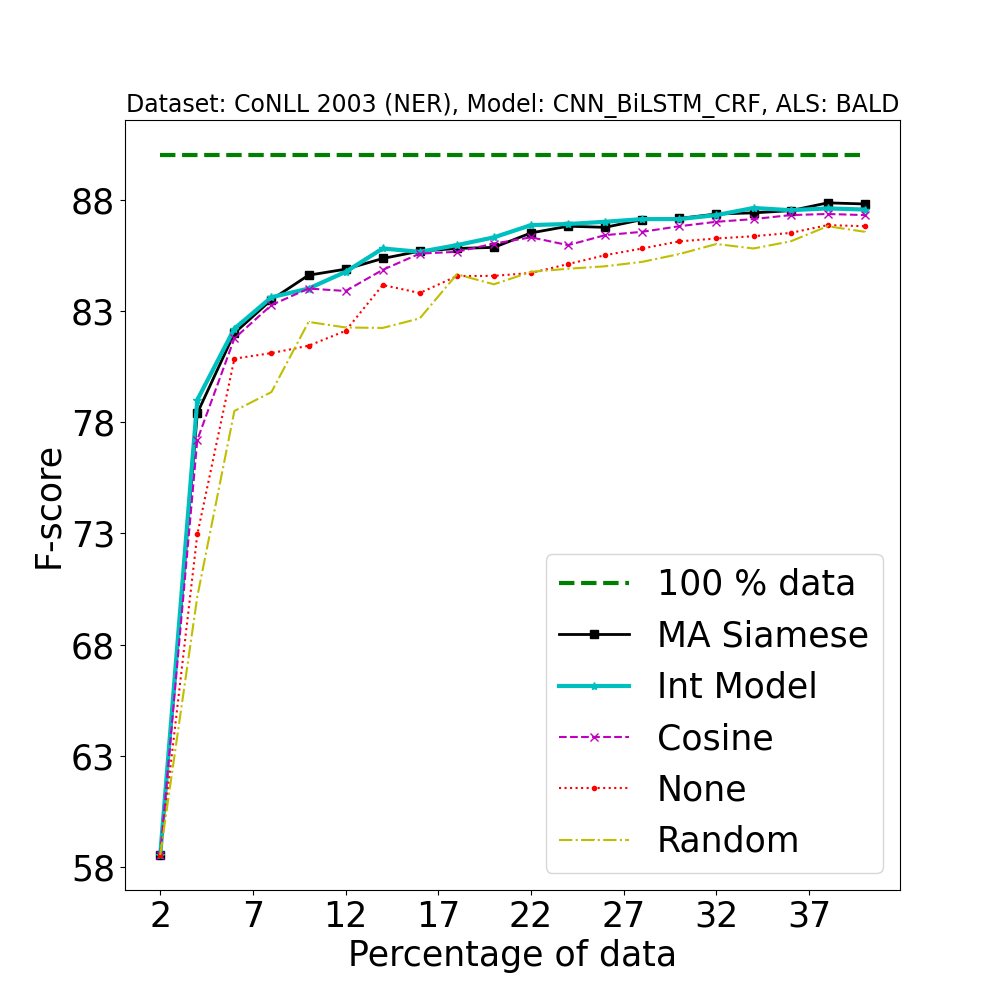}
		% \caption{Plot 4}
	\end{subfigure}%
	~ 
	\begin{subfigure}[t]{0.3\textwidth}
		\centering
		\includegraphics[width=\linewidth]{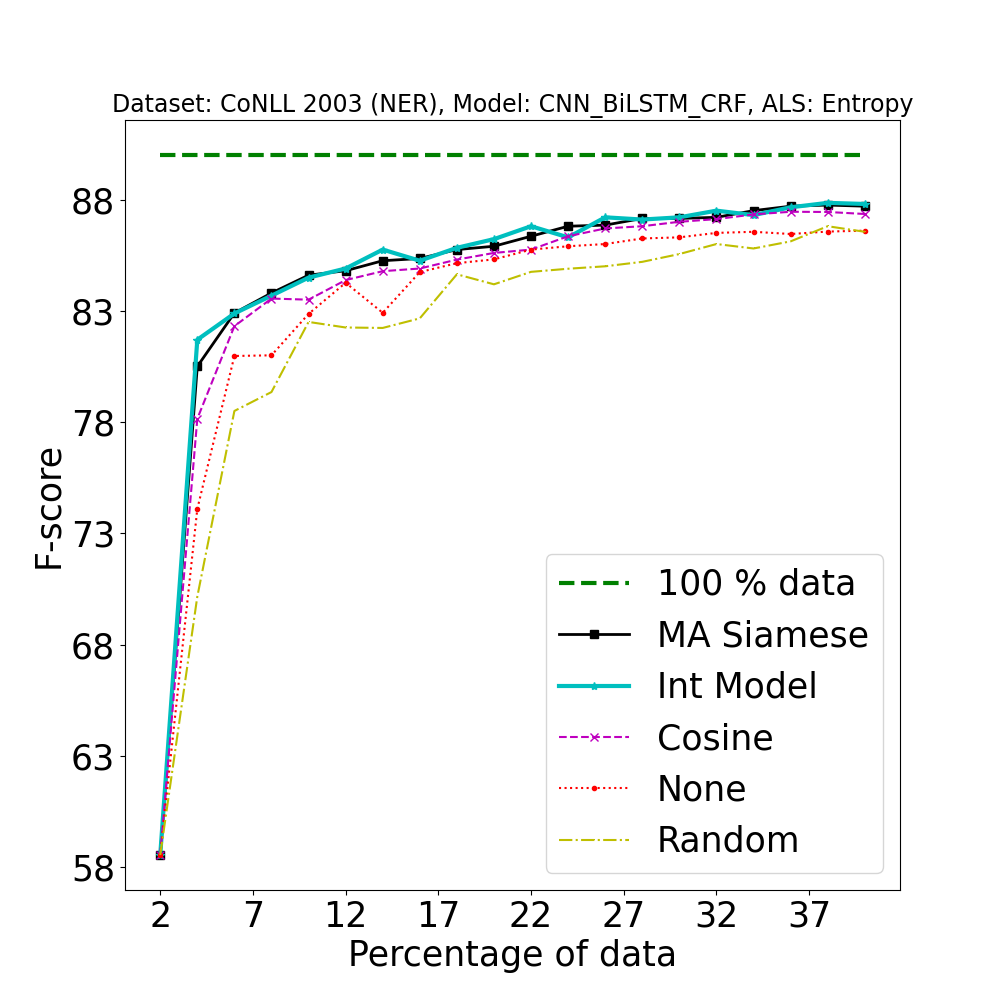}
		%\caption{Plot 5}
	\end{subfigure}%
	~
	\begin{subfigure}[t]{0.3\textwidth}
		\centering
		\includegraphics[width=\linewidth]{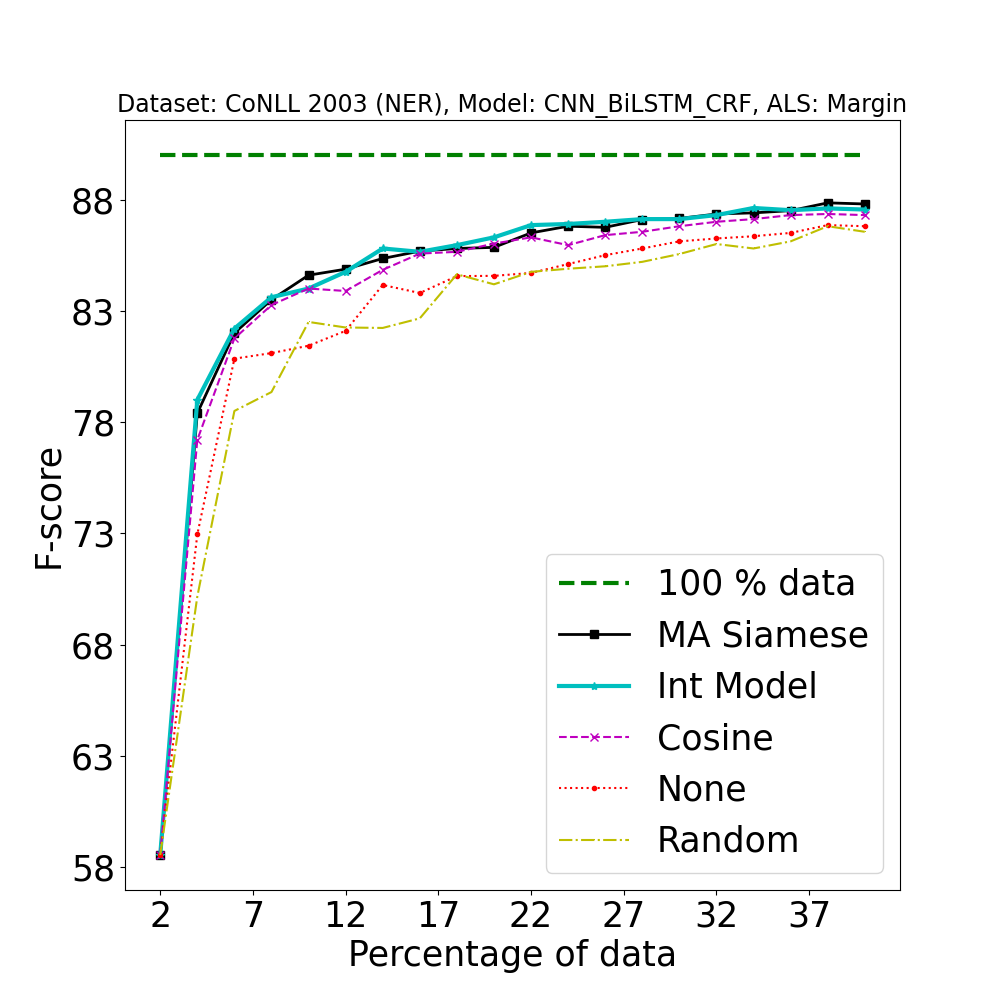}
		%\caption{Plot 6}
	\end{subfigure}%
	\\
	\begin{subfigure}[t]{0.3\textwidth}
		\centering
		\includegraphics[width=\linewidth]{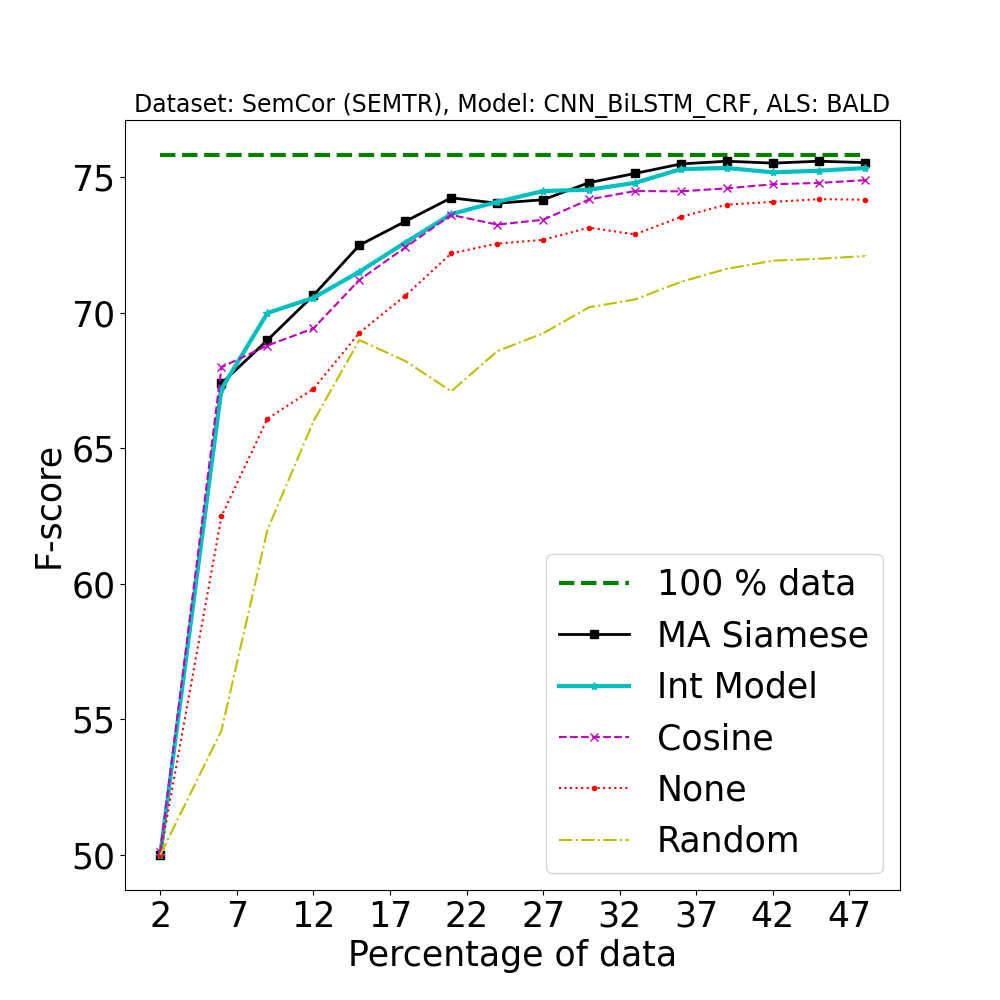}
		%\caption{Plot 1}
	\end{subfigure}%
	~ 
	\begin{subfigure}[t]{0.3\textwidth}
		\centering
		\includegraphics[width=\linewidth]{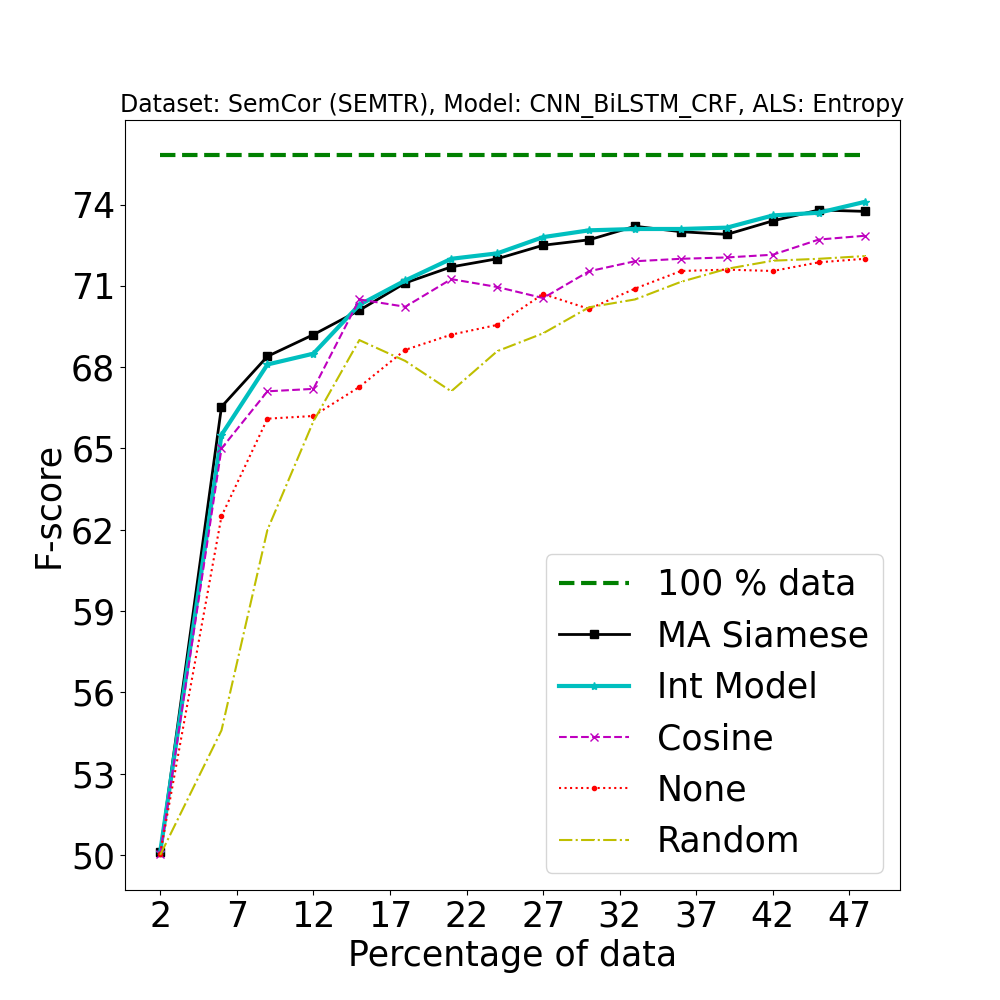}
		%\caption{Plot 2}
	\end{subfigure}%
	~
	\begin{subfigure}[t]{0.3\textwidth}
		\centering
		\includegraphics[width=\linewidth]{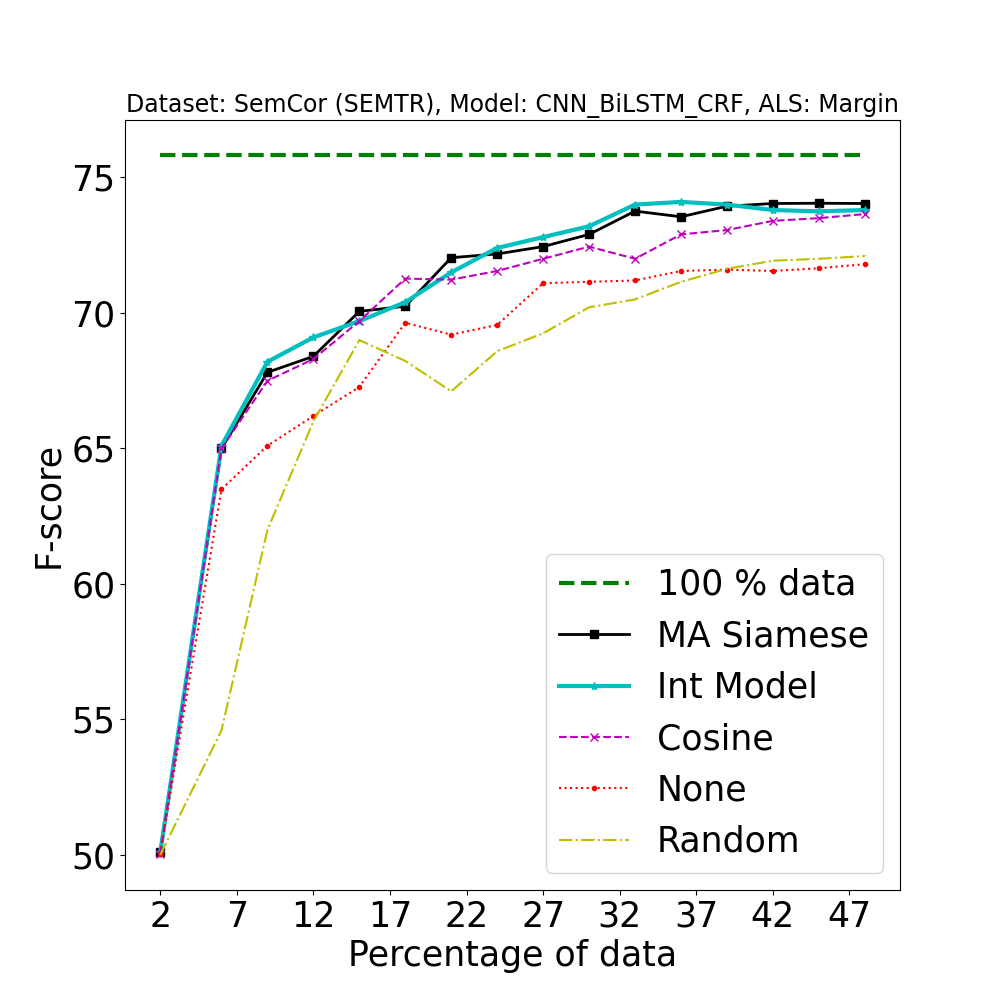}
		%\caption{Plot 3}
	\end{subfigure}%
	\\
	\begin{subfigure}[t]{0.3\textwidth}
		\centering
		\includegraphics[width=\linewidth]{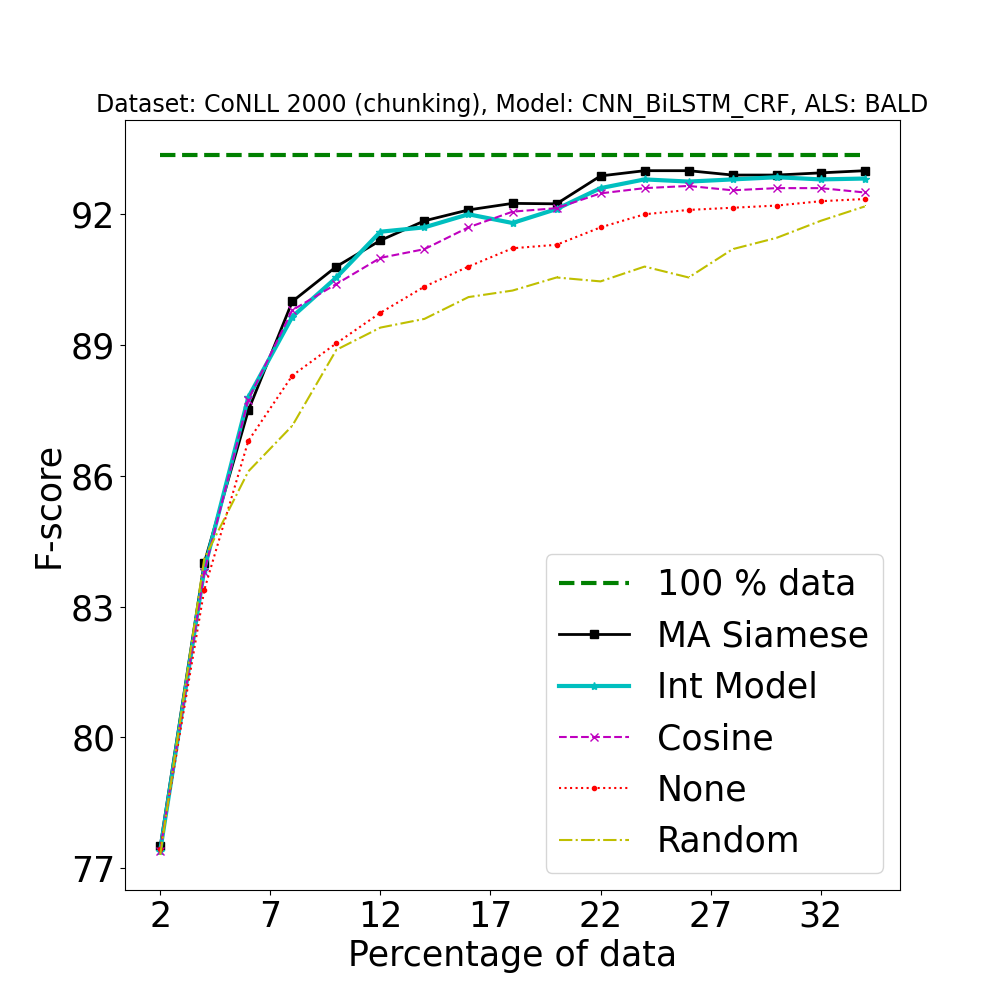}
		%\caption{Plot 10}
	\end{subfigure}%
	~ 
	\begin{subfigure}[t]{0.3\textwidth}
		\centering
		\includegraphics[width=\linewidth]{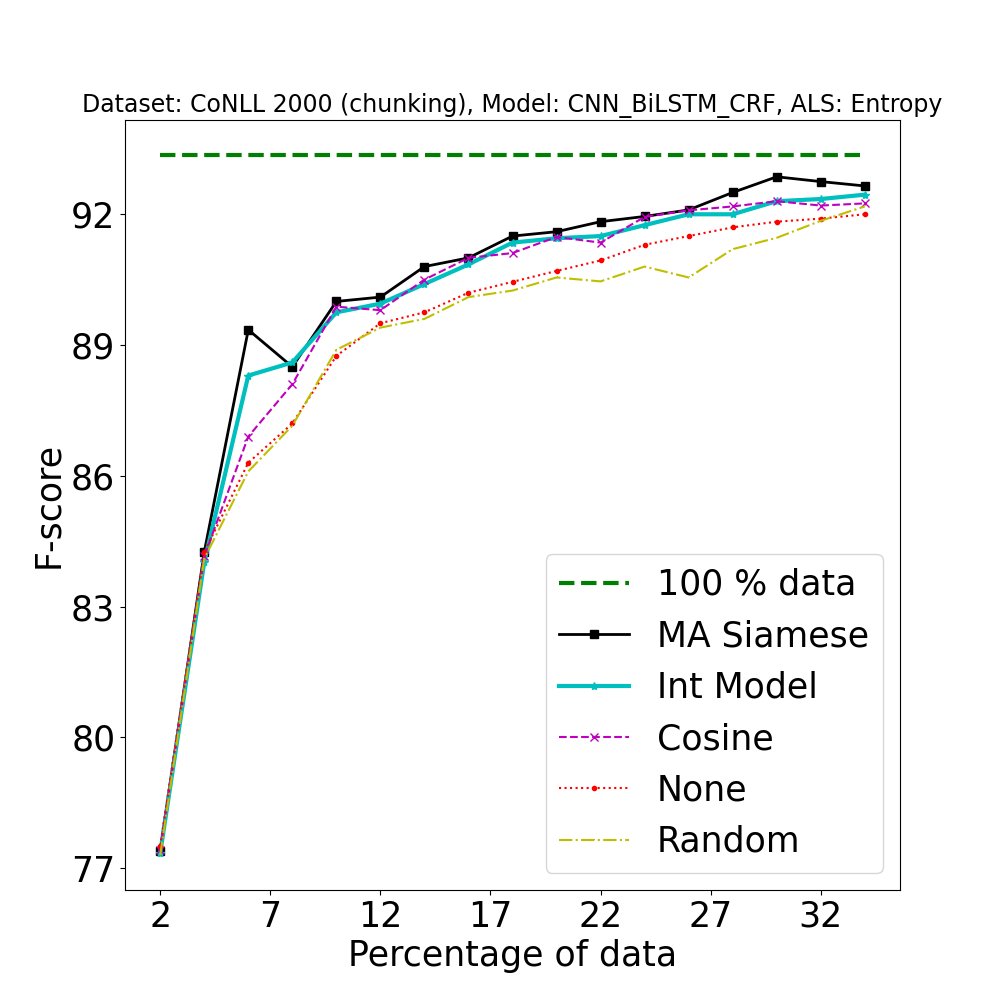}
		%\caption{Plot 11}
	\end{subfigure}%
	~
	\begin{subfigure}[t]{0.3\textwidth}
		\centering
		\includegraphics[width=\linewidth]{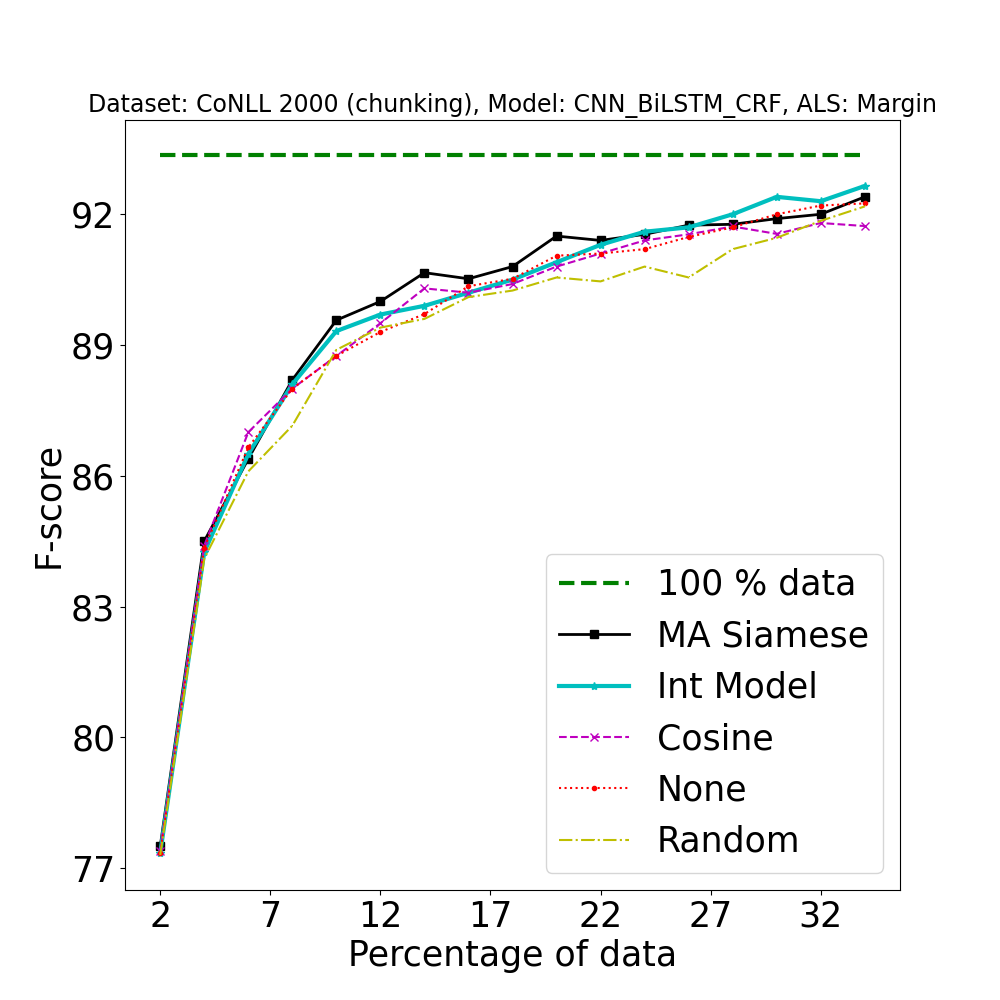}
		%\caption{Plot 12}
	\end{subfigure}
	\caption{[Best viewed in color] Comparison of our approach (A$^\mathbf{2}$L) with baseline approaches on different tasks using different active learning strategies. $1^{st}$ row: POS, $2^{nd}$ row: NER, $3^{rd}$ row: SEMTR, $4^{th}$ row: CHUNK. In each row, from left to right, the three columns represent BALD, Entropy and Margin based AL strategies. Legend Description \{100\% data : full data performance, A$\mathbf{^2}$L (MA Siamese) : Model Aware Siamese, A$\mathbf{^2}$L (Int Model) : Integrated Clustering Model, Cosine : Cosine similarity, None : Active learning strategy without clustering step, Random : Random split (no active learning applied)\}. See Section \ref{section:baselines} for more details. All the results were obtained by averaging over 5 random splits.}
	\label{original_graphs}
\end{figure*}

\begin{table*}[!htbp]
	\small
	\begin{center}
		\begin{tabular}{>{\centering\arraybackslash}m{3.0cm} >{\centering\arraybackslash}m{1.0cm} >{\centering\arraybackslash}m{1.0cm} >{\centering\arraybackslash}m{1.0cm} >{\centering\arraybackslash}m{1.0cm} >{\centering\arraybackslash}m{1.0cm} >{\centering\arraybackslash}m{1.0cm} >{\centering\arraybackslash}m{1.0cm} >{\centering\arraybackslash}m{1.0cm}
		>{\centering\arraybackslash}m{1.0cm}} 
	\toprule 
			\textbf{\diagbox{Setup}{\% data}} & $\mathbf{10\%}$ & $\mathbf{15\%}$ & $\mathbf{20\%}$ & $\mathbf{25\%}$ & $\mathbf{30\%}$ & $\mathbf{35\%}$ & $\mathbf{40\%}$ & $\mathbf{45\%}$ & $\mathbf{50\%}$\\[0.4ex] 
	\midrule
			Iso Siamese & 88.58 & 89.00 & 89.14 & 89.74 & 90.18 & 90.20 & 90.22 & 90.50 & 90.48 \\ 
	\hline
			Cosine & 88.34 & 88.86 & 89.74 & 89.90 & 90.23 & 90.17 & 90.25 & 90.50 & 90.63\\
	\hline
			InferSent & 88.15 & 89.00 & 89.95 & 90.05 & 90.12 & 90.35 & 90.37 & 90.60 & 90.54 \\ 
	\hline
			None (BALD) & 88.58 & 88.50 & 89.23 & 89.51 & 90.00 & 90.05 & 90.12 & 90.40 & 90.44 \\ 
	\hline
			Random (No ALS) & 86.79 & 87.51 & 88.50 & 89.00 & 89.19 & 89.46 & 89.42 & 89.75 & 90.14 \\
	\hline
			\textbf{A$\mathbf{^2}$L (MA Siamese)} & $\mathbf{89.13}$ & 89.50 & 90.10 & $\mathbf{90.34}$ & $\mathbf{90.76}$ & $\mathbf{90.79}$ & $\mathbf{90.80}$ & 90.70 & $\mathbf{90.88}$ \\ 
	\hline
			\textbf{A$\mathbf{^2}$L (Int Model)} & 89.00 & $\mathbf{89.75}$ & $\mathbf{90.20}$ & 90.20 & 90.50 & 90.45 & 90.50 & $\mathbf{90.75}$ & 90.75 \\ 
	\bottomrule
		\end{tabular}
	\end{center}
	\caption{Interpretation of the plot on the top left corner of Fig~\ref{original_graphs} (CoNLL 2003 (POS), BALD) in Appendix. The values in the cells are F-scores on the test set after training on the corresponding percentage of the data. It can be seen that with the increase in \% labeled data, A$\mathbf{^2}$L (MA Siamese) consistently performs better than other baselines.}
	\label{tab_results 2}
\end{table*}

\end{document}